\documentclass[10pt,twocolumn,letterpaper]{article}

\usepackage[pagenumbers]{cvpr} 

\usepackage[utf8]{inputenc} 
\usepackage[T1]{fontenc}    

\usepackage{booktabs}       
\usepackage{nicefrac}       
\usepackage{microtype}      
\usepackage{xcolor}         

\usepackage{epsfig}
\usepackage{algorithm}
\usepackage{algpseudocode}
\usepackage{mathtools}
\usepackage{listings} 
\usepackage{multirow}
\usepackage{ifthen}
\usepackage{adjustbox}
\usepackage{dblfloatfix}
\usepackage{balance}
\usepackage{subcaption}

\usepackage{amsmath,amsfonts}
\usepackage{array}
\usepackage{textcomp}
\usepackage{url}
\usepackage{verbatim}
\usepackage{graphicx}
\usepackage{xspace}
\usepackage{amssymb} 
\usepackage{makecell}

\RequirePackage{color}
\RequirePackage[normalem]{ulem}

\newcommand{\defeq}{\overset{\mathrm{def}}{=\joinrel=}}
\newcommand{\degree}{$^{\circ}$}

\definecolor{flgreen}{HTML}{08FF08}

\definecolor{cvprblue}{rgb}{0.21,0.49,0.74}
\usepackage[pagebackref=true,breaklinks=true,colorlinks=true,bookmarks=false,urlcolor=magenta,citecolor=ForestGreen]{hyperref}

\newcommand{\gitrepo}{\href{https://github.com/rahuldeshmukh43/epimask}{epimask.git}}

\title{EpiMask: Leveraging Epipolar Distance Based Masks in Cross-Attention for Satellite Image Matching}

\author{
    Rahul Deshmukh
    \quad Aditya Chauhan
    \quad Avinash Kak\\[1mm]
    {\tt\small deshmuk5@purdue.edu} 
    \quad {\tt\small chauha35@purdue.edu}
    \quad {\tt\small kak@purdue.edu}\\[1mm]
    Purdue University, West Lafayette \\[2mm]
    \textit{Accepted at CVPR Findings 2026. Official version: } \href{https://openaccess.thecvf.com/content/CVPR2026F/html/Deshmukh_EpiMask_Leveraging_Epipolar_Distance_Based_Masks_in_Cross-Attention_for_Satellite_CVPRF_2026_paper.html}{CVPR-F'26}
}

\begin{document}
\maketitle

\begin{abstract}

The deep-learning based image matching networks can now handle
significantly larger variations in viewpoints and illuminations while
providing matched pairs of pixels with sub-pixel precision.  These
networks have been trained with ground-based image datasets and,
implicitly, their performance is optimized for the pinhole camera
geometry.  Consequently, you get suboptimal performance when such
networks are used to match satellite images since those images are
synthesized as a moving satellite camera records one line at a time of
the points on the ground.  In this paper, we present EpiMask, a
semi-dense image matching network for satellite images that (1)
Incorporates patch-wise affine approximations to the camera modeling
geometry; (2) Uses an epipolar distance-based attention mask to
restrict cross-attention to geometrically plausible regions; and (3)
That fine-tunes a foundational pretrained image encoder for robust
feature extraction. Experiments on the SatDepth dataset demonstrate up
to 30\% improvement in matching accuracy compared to re-trained
ground-based models. The code will be made available through \gitrepo


\end{abstract}

\section{Introduction}
Image matching is the task of finding pixels in two images
that correspond to the same physical point in a 3D scene. It serves as
a fundamental component of 3D reconstruction pipelines, enabling
geometric alignment and 3D reconstruction from multiple
views. Depending on the application, the matching objective can range
from identifying a sparse set of correspondences for image alignment
or a dense set of pixel-level matches for stereo depth
reconstruction. Both paradigms play a critical role in downstream
applications such as structure-from-motion (SfM), multi-view stereo,
and SLAM.

\begin{figure}[!h]
  \centering
  \includegraphics[width=\linewidth]{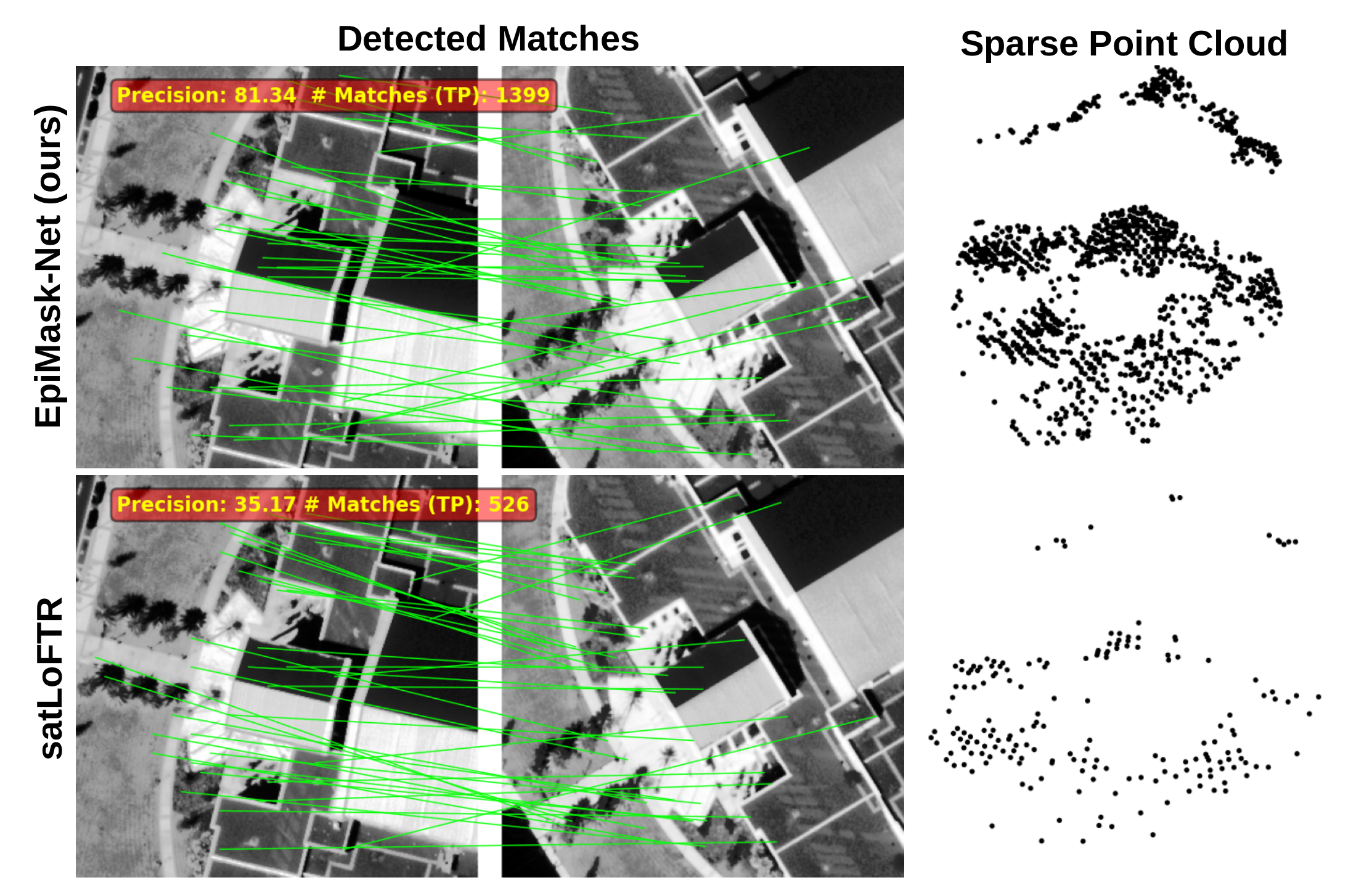}
  %
    \caption{Shown are 40 randomly selected image matches for EpiMask (top row) and SatLoFTR (bottom row).  EpiMask detects more accurate correspondences, as evidenced by a denser and more detailed point cloud on the right.}
  \label{fig:intro_fig}
\end{figure}

The past decade has witnessed an explosion of deep-learning based
approaches that were shown to outperform the classical methods,
starting from simple learnable adaptations of classical methods
\cite{LIFT,KEY_NET, DISK, D2NET} to sophisticated end-to-end trainable
models for detecting sparse
\cite{MagicLeap_CVPR20_SuperGlue,lindenberger2023lightglue},
semi-dense \cite{Sun_CVPR21_LOFTR, Li_NIPS20_DUALRC,
  wang2022matchformer,chen2022aspanformer}, and dense
\cite{edstedt2024roma, edstedt2023dkm} matches. More recently, models
such as RoMa \cite{edstedt2024roma}, LightGlue
\cite{lindenberger2023lightglue}, and GIM \cite{xuelun2024gim}
leverage techniques like large-scale pretraining, using features from
foundational models \cite{oquab2023dinov2}, and fine-tuning to achieve
even higher performance. However, most of these networks are trained
with ground-based imagery and do not explicitly account for the
geometric or photometric characteristics unique to satellite images.

To address challenges specific to satellites imagery, several
satellite image matching datasets \cite{deshmukh2025satdepth,
  whu_stereo, sonali_arxiv_SatStereo} have been proposed, enabling
benchmarking and adaptation of ground-based image matching models to
satellite images. Yet, these efforts typically focus on benchmarking
\cite{song2024deeplearningmeetssatellite} or re-training existing
architectures \cite{deshmukh2025satdepth} on satellite matching
datasets rather than designing models that exploit the rich geometric
metadata available with satellite images. A notable gap in the
literature is that none of the ground-based image matching models has
been tailored to leverage the satellite camera geometry that is always
present in the metadata associated with the satellite images.

\begin{figure}[!h]
    \centering
    \includegraphics[width=\linewidth]{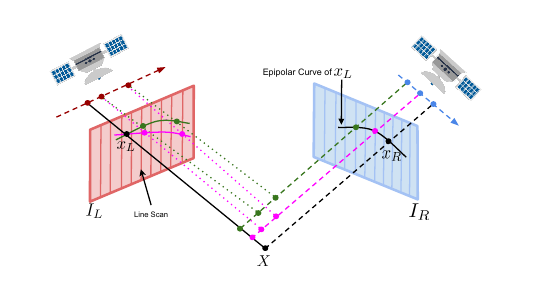}
    \caption{Epipolar geometry of satellite cameras, also known as pushbroom cameras}
    \label{fig:epi_curves}
\end{figure}

A fundamental reason for why the satellite images are more complex
compared to typical ground-based images is that, for the case of
satellites, an image is synthesized one line at a time with a linear
sensor array as the satellite is in motion. A direct manifestation of
this complexity is the fact that the epipolar lines for satellite
images tend to be curved \cite{Franchis_ISPRS14_S2P} as illustrated in
\cref{fig:epi_curves}, whereas for ground-based images (with pinhole
cameras), they tend to be straight lines. For a given pair of images
($I_L, I_R$) for the purpose of matching, the epipolar lines being
curved implies that, for any given pixel $x_L$ in the left image
($I_L$), its corresponding pixel $x_R$ in the right image ($I_R$) will
lie in a region whose shape would be complicated and depend on the
location of $x_L$.  By contrast, with pinhole cameras used for
ground-based imagery, the epipolar lines are straight. Consequently,
the search region for a corresponding pixel is bounded by straight
lines, resulting in simpler downstream logic for image matching.

At this point, a reader may naturally ask: “Why not simply train
existing state-of-the-art image matching networks, originally
developed for ground-based imagery, on satellite images?” After all,
these networks are agnostic to the camera imaging geometry. Their
implicit dependence on the imaging geometry is a consequence of the
datasets on which they are trained. Consider, for instance, the LoFTR
\cite{Sun_CVPR21_LOFTR} network, which achieves remarkable performance
when trained on the MegaDepth \cite{MegaDepthLi18} dataset comprising
images captured with standard pinhole cameras. One could argue that
the pinhole imaging geometry is effectively ``baked'' into the learned
weights. It is therefore reasonable to ask whether re-training such a
model on satellite imagery would yield equally strong performance.

Our work directly investigates the question posed above and arrives at
a surprising conclusion. We show that while a network like LoFTR,
after it is trained with satellite images, performs reasonably well
for image matching tasks, {\em the performance remains suboptimal}.
More specifically, we show that incorporating architectural
modifications that account for the unique characteristics of satellite
imagery, particularly the nonlinear epipolar geometry, yields up to a
30\% improvement in matching accuracy compared to the best that can be
achieved with the original architecture.


The EpiMask architecture proposed in this paper extends LoFTR by
integrating with it a new cross-attention mechanism that directly
addresses the unique epipolar geometry of satellite imagery. The
cross-attention layers are trained using patch-based linear
approximations (modeled through affine cameras) to capture the
non-linear nature of satellite epipolar geometry. Additionally, an
epipolar distance-based attention mask constrains the cross-attention
to geometrically plausible regions. Our model also fine-tunes a
foundational pretrained image encoder to ensure robust feature
extraction. Together, these satellite-specific enhancements to LoFTR
yield a 30\% improvement in image matching performance compared to a
re-trained LoFTR baseline. For a fair comparison, both EpiMask
and LoFTR were trained on the SatDepth dataset
\cite{deshmukh2025satdepth}, designed specifically for satellite
imagery.
\footnote{The name EpiMask denotes the use of an epipolar mask (“EpiMask”) in matching}


As an illustration of the power of EpiMask, the top-left image pair in \cref{fig:intro_fig} shows 40 randomly selected matches detected by our network.  For comparison, the bottom pair shows the same number of matches from satLoFTR trained on the same dataset.  The superior accuracy of matches detected by EpiMask is evident from the higher-quality point cloud shown on the right. A detailed quantitative comparison with competing methods is presented later in the paper.


Our main contributions are: (i) A geometry-aware attention framework for satellite image matching; (ii) Incorporation of a foundational pretrained encoder for improved generalization; (iii) State-of-the-art performance on the SatDepth dataset with up to 30\% improvement; and (iv) Extensive ablation studies validating key design choices.

\section{Related Work}

\textbf{Image Matching Models}: Image matching for ground-based images is an active field of research, so much so that it can be daunting to keep track of all the developments. Broadly speaking, image matching models can be categorized into \textit{sparse}, \textit{semi-dense}, and \textit{dense} matching methods based on the density of correspondences they produce. Sparse methods \cite{MagicLeap_CVPR20_SuperGlue,lindenberger2023lightglue} focus on identifying a limited set of high-confidence keypoint matches for the task of geometric correction. Semi-dense methods \cite{Rocco_NIPS18_NCNET, Rocco_ECCV20_sparseNCNet, Sun_CVPR21_LOFTR, Li_NIPS20_DUALRC, wang2022matchformer,Patch2PixEP,chen2022aspanformer} aim to find correspondences at a downsampled resolution across the entire image, balancing coverage and computational efficiency, for the task of sparse 3D reconstruction. Dense methods \cite{edstedt2024roma, edstedt2023dkm} seek to establish pixel-level correspondences across the entire image to create high-fidelity 3D reconstructions.

The initial design of these models drew inspiration from classical feature-based methods like SIFT \cite{sift, surf, orb} where researchers developed learnable feature detectors and descriptors \cite{LIFT,KEY_NET, D2NET,LF_NET,GEODESC}. However, the field has rapidly evolved towards end-to-end trainable \textit{"detector-free"} architectures \cite{Sun_CVPR21_LOFTR,Li_NIPS20_DUALRC,edstedt2023dkm,Patch2PixEP,wang2022matchformer,edstedt2024roma} that can extract robust features and establish correspondences without explicit keypoint detection or description.

To train these models effectively, various supervision strategies have been employed, such as fully-supervised learning \cite{Sun_CVPR21_LOFTR,Li_NIPS20_DUALRC,wang2022matchformer}, pretraining on synthetic data \cite{MagicLeap_CVPR18_SuperPoint}, and self-supervised learning through homographic adaptation \cite{MagicLeap_CVPR20_SuperGlue,MagicLeap_CVPR18_SuperPoint,lindenberger2023lightglue}. These models were trained on several large datasets such as MegaDepth \cite{MegaDepthLi18}, ScanNet \cite{dai2017scannet}, and many more \cite{aachen_day_night,Vassilieios_CVPR17_HPatches,taira2018inloc}. However, their performance for out-of-distribution scenes remained a challenge. More recent models aim for generalization to diverse scenes by leveraging large-scale pretrained foundation models \cite{edstedt2024roma}, and training on matches mined from internet videos \cite{xuelun2024gim}.

\textbf{Satellite Image Matching}: The satellite image matching literature has taken advantage of the advances made in ground-based image matching. For instance, Ghuffar \etal \cite{ghuffar2022pipeline} uses SuperGlue \cite{MagicLeap_CVPR20_SuperGlue} to automatically match pixels between historical Corona KH-4 images and recent satellite imagery to generate Ground Control Points. Song \etal \cite{song2024deeplearningmeetssatellite} carries out evaluation of pre-trained image matching models for multi-date satellite stereo images. More recently, Deshmukh \etal \cite{deshmukh2025satdepth} introduced the SatDepth dataset for satellite image matching and demonstrated that re-training ground-based image matching models \cite{Sun_CVPR21_LOFTR,wang2022matchformer,Li_NIPS20_DUALRC,Qianqian_ECCV20_CAPS} on SatDepth leads to improved performance.


\textbf{Foundation Model for Satellite Images}:  The \textit{SatlasPretrain} model \cite{bastani2023satlaspretrain} is a large-scale, multi-task foundation model trained on a diverse corpus of satellite imagery comprising over 302M labels across 137 categories and seven annotation types, covering approximately 21M km\textsuperscript{2} of land surface worldwide. Built on a \textit{Swin Transformer}\cite{Liu_2021_ICCV} backbone, it jointly learns from semantic and instance segmentation, object detection, polyline extraction, regression, and classification tasks using spatially aligned multi-temporal image sequences. The model fuses temporal information through max pooling to ensure robustness to illumination, seasonal, and occlusion variations. This broad, heterogeneous, and temporally aligned training enables SatlasPretrain to capture both \textit{fine-grained geometric detail} and \textit{global contextual structure}, producing rich multi-scale and appearance-invariant representations. These characteristics make it an excellent image encoder backbone for feature matching tasks in satellite imagery.

\section{Method}

In this section, we first introduce the fundamentals of satellite
camera and its epipolar geometry. We then describe our masked
cross-attention mechanism and the intuition behind its
design. Finally, we present the details of our model architecture and
implementation specifics.

\subsection{Satellite Camera and Epipolar Geometry}

A satellite image is captured with a pushbroom camera which consists
of a linear array of sensors that records one row of the image at a
time as the satellite moves along its track.  It has been shown that
the complex imaging geometry of satellite sensors can be effectively
approximated by a ratio of two third-degree polynomials. This analytic
form, known as the RPC (Rational Polynomial Coefficients) camera
model, describes the relationship between the image pixel coordinates
and the 3D coordinates of the corresponding points on the ground. For
convenience, we will denote the RPC model by $\mathcal{P}$.  The
nonlinearities in $\mathcal{P}$ cause the curving of the epipolar
lines in satellite image pairs with overlapping views of a scene on
the ground \cite{Franchis_ISPRS14_S2P}.  It has also been shown in the
literature \cite{Franchis_ISPRS14_S2P, okamoto1993orientation} that,
for sufficiently small image patches, the camera model $\mathcal{P}$
can be approximated by an affine camera $\widehat{\mathcal{P}}$. This
property is particularly valuable for applying deep learning
algorithms to satellite images.

\begin{figure}[!h]
    \centering
    \input{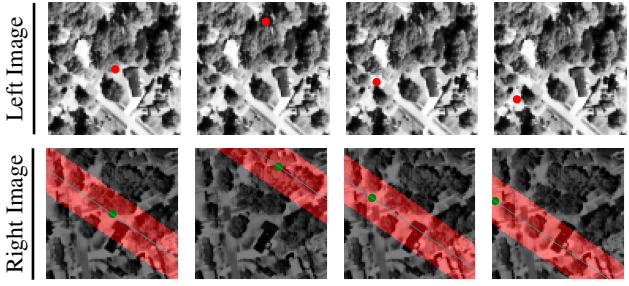}
    \caption{Epipolar distance mask (red band), epipolar line (cyan)
      and corresponding matching point (green dot) in the right image
      patch for a random pixel (red dot) in the left image patch.}
    \label{fig:epi_masks}
\end{figure}

For a pair of satellite image patches ($I_L, I_R$) with affine cameras
($\widehat{\mathcal{P}}_L, \widehat{\mathcal{P}}_R$), we can estimate
the affine fundamental matrix $\mathcal{F}$
~\cite{hartley2003multiple}. Then for the pixel locations
$\boldsymbol{x}_L \in I_L$ and $\boldsymbol{x}_R \in I_R$, we can
compute the symmetric epipolar distance ($d_{sym}$) using
\cref{eq:symmetric_epipolar_distance}. Using a
threshold $\delta_{epi}$, for a pixel $\boldsymbol{x}_L \in I_L$ we
can visualize its corresponding epipolar distance as a banded mask
($\mathcal{M}_{epi}(\boldsymbol{x}_L, \boldsymbol{x}_R) \defeq
d_{sym}(\boldsymbol{x}_L,\boldsymbol{x}_R) < \delta_{epi}$) in $I_R$
shown in \cref{fig:epi_masks}. This banded mask serves as constrained
search region for finding the matching pixel $\boldsymbol{x}_R$ for a
given pixel location $\boldsymbol{x}_L$.

From the RPC model that always accompanies a satellite image as a part
of its metadata, we can readily compute an initial estimate of the
affine fundamental matrix ($\mathcal{F}_0$) and the epipolar mask
$\mathcal{M}_{epi}$ for any given pair of image patches.

\begin{equation}
    d_{sym} =  \frac{0.5\, |(\boldsymbol{x}_L^T\mathcal{F}\boldsymbol{x}_R)|}{ \sqrt{(\mathcal{F}\boldsymbol{x}_L)_1^2 + (\mathcal{F}\boldsymbol{x}_L)_2^2}} + \frac{0.5\, |(\boldsymbol{x}_R^T\mathcal{F}^T\boldsymbol{x}_L)|}{ \sqrt{(\mathcal{F}^T\boldsymbol{x}_R)_1^2 + (\mathcal{F}^T\boldsymbol{x}_R)_2^2}} 
    \label{eq:symmetric_epipolar_distance}
\end{equation}

\subsection{Transformer and Attention Mechanisms}


Transformers \cite{vaswani2017attention} have become a dominant architecture across language, vision, and audio due to their scalability and ability to model long-range dependencies. In vision, their global receptive field enables interactions between distant regions that CNNs struggle to capture. At their core is multi-head attention, derived from scaled dot-product attention (\cref{eq:attention}), which computes similarity scores ($\boldsymbol{S}=\boldsymbol{Q}^\top\boldsymbol{K}$) between queries and keys and uses them to reweight the value features ($\boldsymbol{V}$), producing rich contextual representations.

\begin{equation}
    Attention(\boldsymbol{Q}, \boldsymbol{K}, \boldsymbol{V}) = softmax(\frac{\boldsymbol{\boldsymbol{Q}^T \boldsymbol{K}}}{\sqrt{d_k}}) \boldsymbol{V}
    \label{eq:attention}
\end{equation}

\begin{figure}[!h]
\centering
    \input{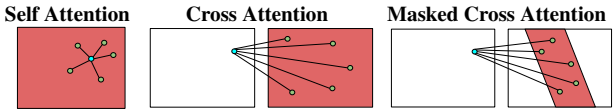}
    \caption{Different attention mechanisms for image matching with
      the query (cyan dot) and keys (green dots) in the valid
      attention region (red region). The masked cross attention uses
      the epipolar distance to attend to the spatially corresponding
      region in the right image patch.}
    \label{fig:attn_mechs}
\end{figure}

Originally developed for machine translation, attention operates in two forms: self-attention, where queries, keys, and values come from the same sequence, and cross-attention, where queries attend to features from another sequence. This paradigm naturally extends to image matching, where a pixel in the left patch ($\boldsymbol{x}_L$) can first refine its representation via self-attention, then use cross-attention to compare against all pixels $\boldsymbol{x}_R \in I_R$ in the right patch to identify correspondences (see \cref{fig:attn_mechs}). Several works \cite{MagicLeap_CVPR20_SuperGlue,Sun_CVPR21_LOFTR,wang2022matchformer} apply these attention mechanisms directly to matching in pinhole camera images.

For patch-based satellite image matching, we now introduce masked
cross-attention (MXA) to enforce the soft geometric constraints using
the initial estimate of the affine fundamental matrix
($\mathcal{F}_0$). The MXA mechanism uses an attention mask
$\mathcal{M}_{epi}(\boldsymbol{x}_L, \boldsymbol{x}_R)$. This mask
restricts the cross-attention of each pixel $\boldsymbol{x}_L$ in the
left image patch to a spatially corresponding region along its
corresponding epipolar band in the right image $I_R$ patch, thereby
enforcing geometric consistency during feature matching, as
illustrated in \cref{fig:attn_mechs}.

\begin{figure*}[ht]
    \centering
    \includegraphics[width=\linewidth]{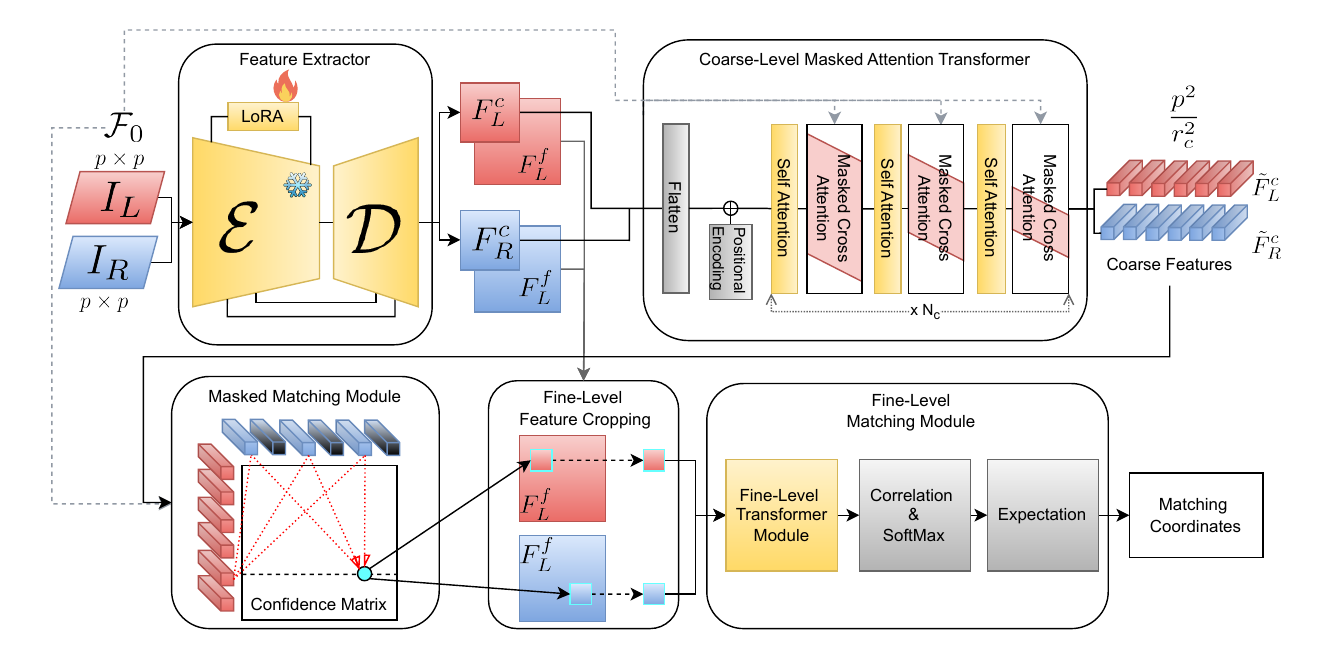}
    \caption{ Network Architecture for EpiMask: Given a pair of
      satellite image patches ($I_L, I_R$) and the approximate affine
      fundamental matrix ($\mathcal{F}_0$), coarse ($F_L^c, F_R^c$)
      and fine ($F_L^f, F_R^f$) features are extracted using a
      pretrained and lora-finetuned encoder-decoder network. The
      coarse features are refined to ($\tilde{F}_L^c, \tilde{F}_R^c$)
      using self and epipolar masked cross attention transformer
      layers. The cross attention layers use $\mathcal{F}_0$ for
      attention masking with decreasing band width (red band) across
      the layers. Coarse features ($\tilde{F}_L^c, \tilde{F}_R^c$)
      with attention mask are used to identify coarse matches
      $\mathcal{M}_c = \{\tilde{i}, \tilde{j}\}$. The coarse matches
      are used to crop the fine-level features that are subsequently
      processed by the fine-level transformer for refining the
      features. The transformed features are used to identify the
      fine-level match coordinates $\mathcal{M}_f = \{\hat{i},
      \hat{j}\}$ using a correlation and softmax based probability
      score followed by the expectation operator to get the matching coordinate $\hat{j}
      \in I_R$ corresponding to the cropped-patch-center coordinate $\hat{i}
      \in I_L$.}
    \label{fig:network_arch}
\end{figure*}

\subsection{Model Architecture}
When given a pair of left and right $p \times p$ sized satellite image
patches ($I_L, I_R$) along with the initial estimate of affine
fundamental matrix ($\mathcal{F}_0$), our model detects matching
points by focusing on the epipolar bands. Our model architecture as
shown in \cref{fig:network_arch} is derived from LoFTR
\cite{Sun_CVPR21_LOFTR} with the following two key changes in order to
adapt the model to satellite image matching: (1) We change the image
encoder to a satellite specific image encoder; and (2) We improve the
coarse-level transformer and matching module with masked attention. We
present further details below.

\subsubsection*{Feature Extractor}\label{sec:feature_extractor}

We extract coarse $(F_L^c, F_R^c)$ and fine $(F_L^f, F_R^f)$ features
from the patch pair $(I_L, I_R)$ using an FPN-style encoder-decoder
\cite{lin2017feature_FPN}. The encoder $\mathcal{E}$ is initialized
from the frozen Satlas-Pretrain model \cite{bastani2023satlaspretrain}
and equipped with randomly initialized learnable LoRA layers
\cite{hu2022lora}. The decoder $\mathcal{D}$ employs bilinear
upsampling followed by skip connection fusion and convolutional
refinement. Coarse and fine features are produced at spatial
resolutions $(p/r_c, p/r_f)$ with $(r_c, r_f) \in \{(4, 2), (8, 2)\}$.
We have evaluated two skip-fusion strategies: (1) element-wise
addition, and (2) concatenation followed by convolution.

\subsubsection*{Coarse-Level Masked Attention Transformer}

After feature extraction the coarse features are flattened and added
to 2D sinusoidal positional encodings
\cite{carion2020end_detr_pos_embedding,Sun_CVPR21_LOFTR}. The features
are then passed through a masked attention transformer resulting in
transformed coarse features ($\tilde{F}_L^c, \tilde{F}_R^c$). The
masked attention transformer is composed of $N_{c}$ interleaved self
and masked cross attention layers. The self attention layer uses
linear attention formulation as per
\cite{katharopoulos2020transformers_linearattention,
  Sun_CVPR21_LOFTR}. The masked cross attention layers follow the
architecture of transformer \cite{vaswani2017attention} and uses the
epipolar mask $\mathcal{M}_{epi}$ as the attention mask. The mask
restricts attention to pixels within an epipolar band of width $b =
2\delta_{epi}$. To gradually introduce the geometric constraints, the
band width is linearly decreased from $b=p$ to $b=\gamma p$ across the
masked attention layers as shown in
\cref{fig:network_arch}.


\subsubsection*{Masked Coarse Matching Module}
Previous works have used either a dual-softmax operation \cite{Rocco_NIPS18_NCNET, Sun_CVPR21_LOFTR} or an optimal transport layer \cite{MagicLeap_CVPR20_SuperGlue, Sun_CVPR21_LOFTR} as differentiable matching modules. In our model, we adopt the dual-softmax operator and extend it to incorporate the attention mask. 
Specifically, we first compute a similarity matrix between the transformed coarse features then modulate the similarity by the attention mask $\mathcal{M}_{epi}$ to obtain a masked matching confidence matrix $P_c$, as defined in \cref{eq:masked_coarse_matching}. We use the attention mask from the final coarse transformer layer, with a band width of $b=\gamma p$, as the mask for the matching layer.
\begin{equation}
    \begin{aligned}
    \tilde{S}(i,j) &= \begin{cases} \frac{<\tilde{F}_L^c(i) , \tilde{F}_R^c(j)>}{\tau} & \text{if } \mathcal{M}_{epi}(i,j) \text{ is True}  \\ -\infty & \text{otherwise} \end{cases} \\
    P_c &= softmax(\tilde{S}(i, \cdot))_j \odot softmax(\tilde{S}(\cdot, j))_i
    \end{aligned}
    \label{eq:masked_coarse_matching}
\end{equation}

Then, following \cite{Sun_CVPR21_LOFTR}, we select a set of matches
$\mathcal{M}_c$ using a confidence threshold $\delta_c$ on the
confidence matrix $P_c$, and enforce the mutual nearest neighbor (MNN)
criterion. The coarse-level match prediction set is given by
$\mathcal{M}_c = \{(\tilde{i}, \tilde{j})| \forall (\tilde{i},
\tilde{j}) \in \text{MNN} (P_c), P_c(\tilde{i}, \tilde{j}) \geq
\delta_c \}$.

\subsubsection*{Fine-Level Matching Module}
Using the coarse correspondences, we refine the match locations to pixel-level precision through a coarse-to-fine refinement module following LoFTR \cite{Sun_CVPR21_LOFTR}. The module extracts $w \times w$ feature crops from the fine-level feature maps $(F_L^f, F_R^f)$ centered at each coarse match $(\tilde{i}, \tilde{j})$. These cropped features are then passed through the LoFTR transformer to obtain locally refined feature representations. Since the crop size is significantly smaller than the epipolar mask bandwidth ($w \ll b$), we omit the epipolar masking at this stage. The refined features are used to compute a correlation score between the center feature of the left crop and all features in the corresponding right crop. This correlation score is normalized into a probability distribution, and the final sub-pixel matching coordinates $\mathcal{M}_f = {(\hat{i}, \hat{j})}$ are obtained using the expectation operator following \cite{Sun_CVPR21_LOFTR, Qianqian_ECCV20_CAPS}.

\subsubsection*{Training Loss}
We train the model with full supervision using the, $L = L_c + L_f$, coarse-level and fine-level loss formulation of LoFTR \cite{Sun_CVPR21_LOFTR} as shown in \cref{eq:losses}. The coarse-level loss function ($L_c$) is the cross-entropy loss over the predicted coarse confidence matrix $P_c$ and the fine-level loss function ($L_f$) is the Mean Squared Error weighted by the inverse of the variance for the predicted coordinate $\hat{j}$ and the true coordinate $\hat{j}_{gt}$.

For supervision, we follow SatDepth \cite{deshmukh2025satdepth} and
use the affine cameras ($\widehat{\mathcal{P}}_L,
\widehat{\mathcal{P}}_R$) with corresponding SatDepth maps to warp
points from one image patch to the other and then carry out 3D
distance checks for identifying ground truth matches. This involves
four steps - (1) Create coarse grid of points ($\{\tilde{i}\}$) for
$I_L$ and use the SatDepth Map for $I_L$ to get the world points
$\{X_L\}$ (2) Warp the world points to $I_R$ using the right camera
$\tilde{j} = \widehat{P}_R(X_L)$ (3) Using the SatDepth maps for
$I_R$, get the coordinates of the world point $X_R$ visible in $I_R$
at $\tilde{j}$ (4) Carry out the distance check using $||X_L -
X_R||_2^2<\delta_{3D}$. To compute the final ground-truth labels for
the coarse loss, we calculate mutually nearest neighboring coarse
matches by warping points from left-to-right and right-to-left and the
set of points which satisfy both directions is the set of ground-truth
matches $\mathcal{M}_c^{gt}$.

\begin{equation}
    \begin{aligned}
L_c &= -\frac{1}{|\mathcal{M}_{c}^{gt}|} \sum_{(\tilde{i}, \tilde{j})\in \mathcal{M}_c^{gt}} log \, \mathcal{P}_c(\tilde{i}, \tilde{j})\\
L_f &= \frac{1}{|\mathcal{M}_f|} \sum_{(\hat{i}, \hat{j})\in \mathcal{M}_f}\frac{1}{\sigma^2(\hat{i})}||\hat{j} - \hat{j}_{gt}||_2^2
    \end{aligned}
\label{eq:losses}
\end{equation}

\subsection{Implementation Details}

\subsubsection*{Training Data} 
We train and evaluate our model using the SatDepth dataset
\cite{deshmukh2025satdepth} along with its rotation augmentation
strategy. All experiments are conducted using the official train,
validation, and test splits.  SatDepth provides approximately 12.8k
training image pairs. 
However, due to its online patch-sampling training strategy, the data pipeline can become bottlenecked by disk I/O, causing individual experiments to run for several days.
To mitigate this, we precompute and store 128k (10x) randomly sampled
training pairs as a sharded WebDataset, with each shard containing 500
samples. During training, we randomly select 48 shards per epoch,
ensuring diverse patch combinations across epochs and effectively
mimicking SatDepth's original sampling strategy. This optimization
substantially reduces disk I/O overhead, decreasing the total training
time from around five days to about one day. We share the scripts for
generating the sharded WebDataset from the original SatDepth dataset.

\subsubsection*{Architectural Details} 
We use the `Aerial-SwinB-SI' pretrained checkpoint from Satlas as our image encoder. The Satlas encoder generates feature maps up to a spatial resolution of $1/32$ of the input image; however, in our feature extractor, we utilize feature maps only up to the $1/16$ resolution level. Since the Satlas encoder expects a three-channel input, following SatDepth \cite{deshmukh2025satdepth}, we replicate the single-channel grayscale satellite image three times along the channel dimension to ensure compatibility. The Swin architecture employs a combined linear layer for the query, key, and value projections. When applying LoRA to this layer, we use a LoRA rank that is three times larger to account for the shared projection structure. Further architectural and implementation details are provided in the supplementary material.

\subsubsection*{Training Details} 
We train our model for 30 epochs using a multi-step learning rate scheduler and Adam-W optimizer, following the training setup of SatDepth \cite{deshmukh2025satdepth}. For attention masking, we adopt a warm-up strategy where no masking is applied during the first $N_m = 5$ epochs, after which a linearly decreasing epipolar band width is applied across the masked attention layers. The training is performed in two stages: in the first stage, we train the model without LoRA layers to obtain stable weights for the decoder, the coarse-level and fine-level transformer modules. In the second stage, we initialize the model with the stage-one weights and introduce learnable LoRA layers in the encoder for fine-tuning. 
We follow the evaluation metrics and protocol of SatDepth for all experiments, except that we set the number of top matches to K=2000 (SatDepth uses K=200) since our model produces a larger number of reliable correspondences.
All experiments were conducted on 4x NVIDIA RTX A6000 GPUs with DDP and gradient accumulation.

\section{Results}

\begin{figure*}[!ht]
    \centering
    \includegraphics[width=\linewidth]{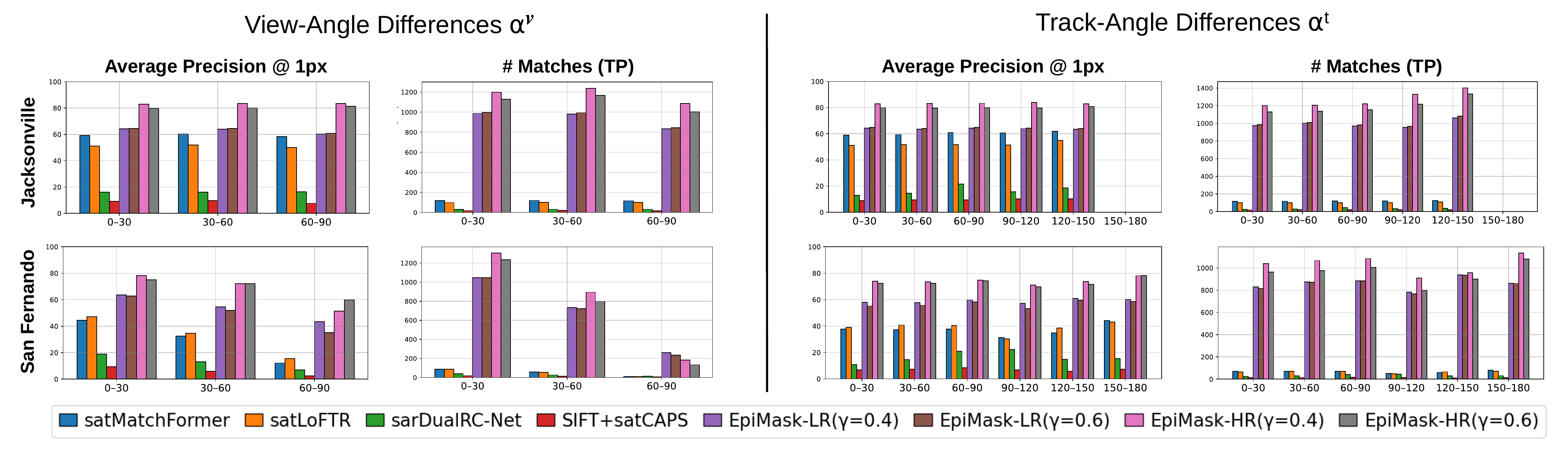}
    \caption{Comparison of all model average precision and number of true-positive matches for Jacksonville and San Fernando Testing AOIs of SatDepth in the simulated rotation experiment. EpiMask consistently achieves highest precision and detects the largest number of matches w.r.t varying view-angle differences $\alpha^{v}$ and track-angle differences $\alpha^{t}$.}
    \label{fig:all_model_comparison_plots}
\end{figure*}

\begin{figure*}
    \centering
    \input{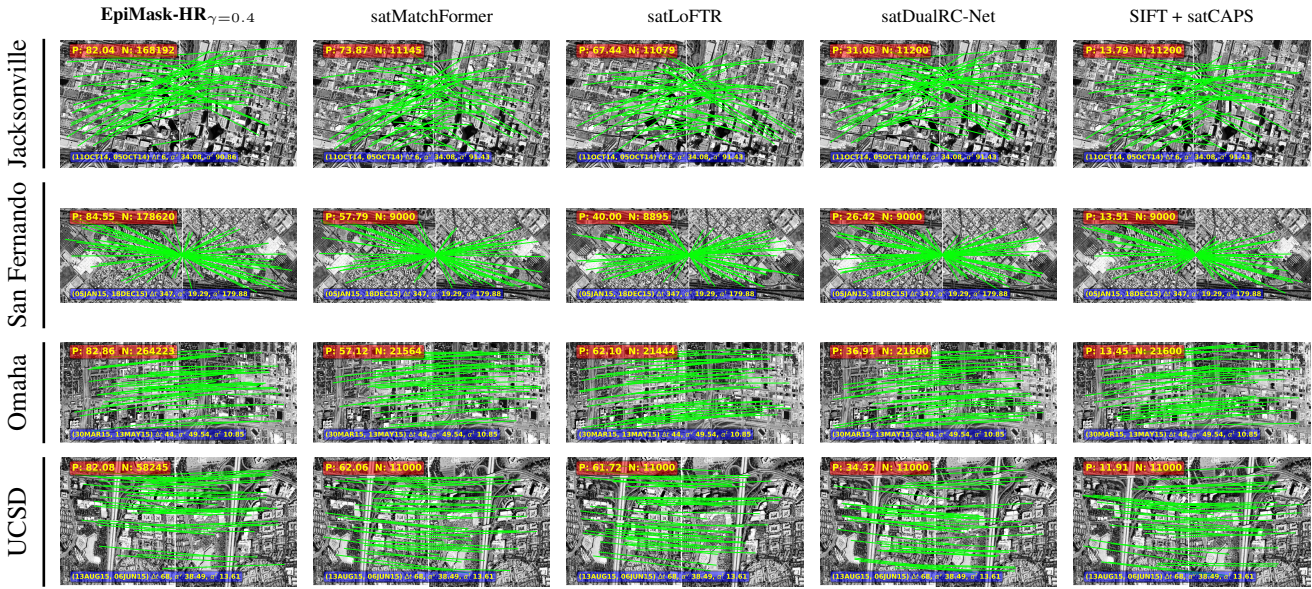}
    \caption{Qualitative comparison of our model results against other models for large track-angle difference ($\alpha^t$) -- our model has the highest precision score. Precision (P) and number of matches (N) are displayed at the top of each plot. Image pair names, time difference ($\Delta t$), view-angle difference ($\alpha^v$), and track-angle difference ($\alpha^t$) are displayed at the bottom. The green lines depict 40 randomly chosen true matches.}
    \label{fig:qualitative_matching_plots_rel_track}
\end{figure*}

We have evaluated the proposed EpiMask with four configurations
for (1) Combining the coarse-fine resolution (High-Res (HR) with
$r_c{=}4,r_f{=}2$ and Low-Res (LR) with $r_c{=}8,r_f{=}2$); and (2) Epipolar
mask band widths ($\gamma{=}0.4$ and $\gamma{=}0.6$). All models are
trained and evaluated following the SatDepth protocol, with
comparisons made against baseline presented in SatDepth. We present
quantitative and qualitative results for SatDepth testing AOIs in
\cref{tab:matcher_performance_main,fig:qualitative_matching_plots_rel_track},
respectively. Following SatDepth~\cite{deshmukh2025satdepth}, we also
present quantitative results for `Simulated-Rotation' on all testing
AOIs in \cref{fig:all_model_comparison_plots}. For the aggregated
metrics in \cref{tab:matcher_performance_main}, we weight each angular
bin by the inverse of its sample count to correct for track-angle
imbalance, following \cite{deshmukh2025satdepth}.

\begin{table}[!h]
    \centering
    \resizebox{\linewidth}{!}{%
    \setlength\tabcolsep{1.5pt}
    \begin{tabular}{ccccccc}
    \toprule
    \multicolumn{7}{c}{\centering \textbf{Jacksonville} \textbf{|} \textbf{San Fernando}}\\
    \midrule
    &\multirow{2}{*}[-.4em]{Method} 
    & \multicolumn{3}{c}{Pose estimation AUC $\uparrow$} 
    & \multirow{2}{*}[-.4em]{\begin{tabular}[c]{@{}c@{}}Precision $\uparrow$\\ @1px \end{tabular}} 
    & \multirow{2}{*}[-.4em]{\begin{tabular}[c]{@{}c@{}} \# Matches $\uparrow$\\ ({\scriptsize{TP}})\end{tabular}}\\
    \cmidrule(lr){3-5}
    & &@5\degree & @10\degree & @20\degree & &\\
    \midrule
    \multirow{4}{*}{\rotatebox[origin=c]{90}{SatDepth\cite{deshmukh2025satdepth}}}
    &\small{SIFT + satCAPS}~\cite{Qianqian_ECCV20_CAPS}  & 38.49 \textbf{|} 36.75 & 43.26 \textbf{|} 40.09 & 50.94 \textbf{|} 45.69 & 10.67 \textbf{|} 7.89 & 21 \textbf{|} 16 \\
    &\small{satDualRC-Net}~\cite{Li_NIPS20_DUALRC}  & 41.19 \textbf{|} 40.57 & 47.57 \textbf{|} 46.77 & 56.31 \textbf{|} 55.58 & 19.94 \textbf{|} 15.88 & 40 \textbf{|} 32 \\
    &\small{satLoFTR}~\cite{Sun_CVPR21_LOFTR}  & 78.48 \textbf{|} 53.60 & 87.02 \textbf{|} 62.96 & 92.30 \textbf{|} 71.34 & 54.87 \textbf{|} 42.58 & 108 \textbf{|} 71 \\
    &\small{satMatchFormer}~\cite{wang2022matchformer}  & 81.37 \textbf{|} 54.57 & 89.01 \textbf{|} 64.15 & 93.56 \textbf{|} 72.68 & 61.96 \textbf{|} 39.83 & 124 \textbf{|} 73 \\
    \cmidrule{2-7}
    \multirow{4}{*}{\rotatebox[origin=c]{90}{\textbf{Ours}}}
    &\small{EpiMask-LR$_{\gamma=0.6}$}  & 89.32 \textbf{|} 79.62 & 93.67 \textbf{|} 87.92 & 96.13 \textbf{|} 93.11 & 62.38 \textbf{|} 39.23 & 1027 \textbf{|} 134 \\
    &\small{EpiMask-LR$_{\gamma=0.4}$}  & 89.51 \textbf{|} 80.38 & 93.69 \textbf{|} 88.58 & 96.06 \textbf{|} 93.52 & 61.95 \textbf{|} 45.60 & 1007 \textbf{|} \textbf{148} \\
    &\small{EpiMask-HR$_{\gamma=0.6}$}  & 91.53 \textbf{|} 85.92 & 94.81 \textbf{|} 91.08 & 96.66 \textbf{|} 94.17 & 81.29 \textbf{|} 72.78 & 1187 \textbf{|} 56 \\
    &\small{EpiMask-HR$_{\gamma=0.4}$}  & \textbf{92.66} \textbf{|} \textbf{87.52} & \textbf{95.57} \textbf{|} \textbf{92.73} & \textbf{97.14} \textbf{|} \textbf{95.80} & \textbf{83.32} \textbf{|} \textbf{70.57} & \textbf{1286} \textbf{|} 113 \\
    \bottomrule
    \end{tabular}
    }
    \caption{Weighted average of Precision, Pose error, and number of True Positive (TP) matches over all testing image patches for Jacksonville and San Fernando AOIs.}
    \label{tab:matcher_performance_main}
\end{table}

From \cref{tab:matcher_performance_main}, we observe that (1) all EpiMask variants outperform the baseline models from \cite{deshmukh2025satdepth}, (2) the High-Res configurations produce more accurate and denser correspondences than the Low-Res ones, (3) the model achieves approximately 90\% pose-estimation accuracy, making it a perfect choice for image alignment task, and (4) The model extracts large number of matches  making it suitable for generating sparse point clouds.

Moreover, the simulated-rotation results in \cref{fig:all_model_comparison_plots} show that all EpiMask variants consistently outperform baselines across all ranges of view-angles ($\alpha^{v}$) and track-angles ($\alpha^{t}$). Performance degrades only under extreme view-angle differences, particularly for unseen AOIs, where such configurations are under-represented in training.

We present key ablation studies below, with additional results, angle-wise metrics, and training loss plots provided in the supplementary material.

\subsection{Resolution Ablation}
\begin{figure}[H]
    \centering
    \input{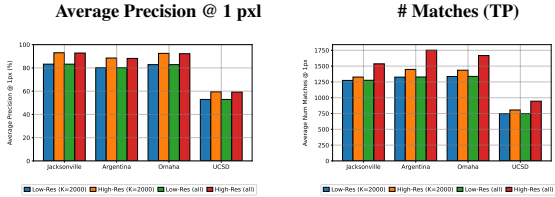}
    \vspace*{-1ex}
    \caption{Average precision and number of true-positive matches for all testing AOIs for High-Res ($r_c=4, r_f=2$) and Low-Res ($r_c=8, r_f=2$) resolution configurations.}
    \label{fig:ablation_performance_res}
\end{figure}

We evaluate the impact of spatial resolutions of coarse and fine features on model performance for two settings - (1) High-Res ($r_c=4, r_f=2$), and (2) Low-Res ($r_c=8, r_f=2$). We evaluate these models for top (K=2000) matches as well as for all detected matches $\mathcal{M}_f$. As can be seen from \cref{fig:ablation_performance_res}, the High-Res setting performs the best.

\vspace*{-1ex}

\subsection{Attention Mask Width Ablation}
\begin{figure}[H]
    \centering
    \input{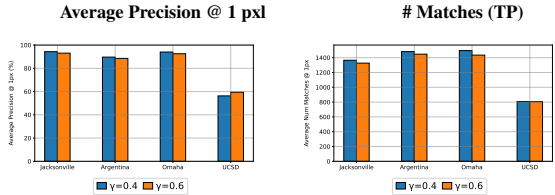}
    \vspace*{-1ex}
    \caption{Average precision and number of true-positive matches for all testing AOIs for models with different attention mask widths ($\gamma=0.4$ and $\gamma=0.6$).}
    \label{fig:ablation_performance_mask_width}
\end{figure}

We evaluate the effect of varying the attention mask width in our Coarse-Level Masked Attention Transformer, comparing $\gamma=0.4$ and $\gamma=0.6$. As shown in \cref{fig:ablation_performance_mask_width}, performance remains largely unchanged, indicating that the model inherently focuses on geometrically consistent regions within the epipolar band. This robustness suggests that fine-grained tuning of mask width is unnecessary. In practice, narrower masks may be preferred for newer satellites with accurate pose metadata, while wider masks can better handle older sensors with noisier estimates of camera pose.
\vspace*{-1ex}

\subsection{Positional Encoding Ablation}
\begin{figure}[H]
    \centering
    \input{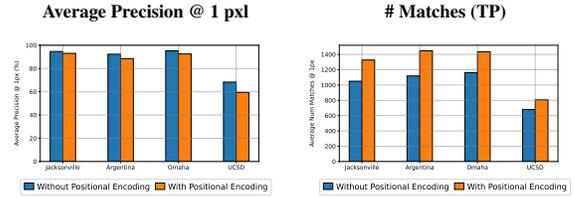}
    \vspace*{-1ex}
    \caption{Average precision and number of true-positive matches for all testing AOIs for models with and without positional encoding.}
    \label{fig:ablation_performance_pos_enc}
\end{figure}

We assess the impact of positional encoding in the coarse-level masked attention transformer. As shown in \cref{fig:ablation_performance_pos_enc}, average precision remains similar, but positional encodings consistently increase true-positive matches.

\vspace*{-1ex}

\subsection{Fine-Tuning Ablation}
\begin{figure}[H]
    \centering
    \input{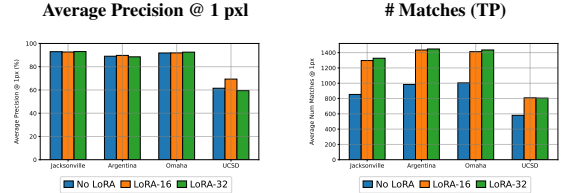}
    \vspace*{-1ex}
    \caption{Average precision and number of true-positive matches for all testing AOIs for models with different fine-tuning strategies (No Fine-tuning, LoRA-16, LoRA-32).
    }
    \label{fig:ablation_performance_lora}
\end{figure}

We evaluate the effect of fine-tuning the image encoder using LoRA with two ranks: 16 (LoRA-16) and 32 (LoRA-32). Both are compared against the baseline model trained without LoRA. As shown in \cref{fig:ablation_performance_lora}, LoRA significantly increases the number of true positives, indicating that lightweight fine-tuning effectively adapts the pretrained backbone to our matching task. Increasing the rank from 16 to 32 provides negligible gains, suggesting that a moderate rank of 16 strikes a good balance between parameter efficiency and adaptation quality.

\vspace*{-1ex}

\subsection{Feature Extractor Ablation}
\begin{figure}[H]
    \centering
    \input{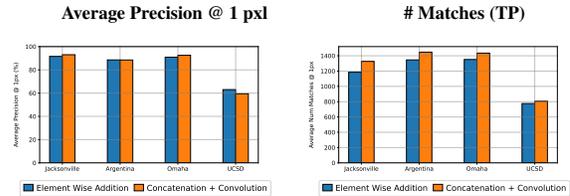}
    \caption{Average precision and number of true-positive matches for all testing AOIs for models with different feature extractor skip-fusion strategies.
    }
    \label{fig:ablation_performance_fe}
\end{figure}

As discussed in \cref{sec:feature_extractor}, we experimented with two different strategies for skip-fusion in our Feature Extractor. As shown in \cref{fig:ablation_performance_fe}, the concatenation followed by convolution strategy performs better compared to na\"{\i}ve element-wise addition.

\section{Discussion}


In this work, we presented EpiMask, a new image matching
framework meant for semi-dense correspondence estimation in satellite
images. Our approach fine-tunes a foundational pretrained encoder
using a novel epipolar-distance based masked cross-attention module,
which enables the network to restrict attention to geometrically valid
regions and to thus produce accurate and highly generalizable
matches. EpiMask achieves state-of-the-art performance on the
SatDepth dataset, and our ablations highlight the architectural
components that contribute the most to the performance gains. We
believe EpiMask provides immense value to the remote sensing
community, enabling robust image matching for the task of image
alignment and sparse 3D reconstruction. A limitation of our study is
the lack of evaluation across globally diverse geographic regions,
primarily due to the scarcity of suitable datasets.

Finally, although our experiments focus on satellite images, the
proposed masked cross-attention mechanism is broadly applicable to any
domain with known camera metadata — for example, UAV imaging with
onboard pose sensors or multi-view X-ray systems (e.g., angiography)
with calibrated image acquisition geometry.

\clearpage  

\onecolumn  
\begin{center}
    \textbf{\LARGE Supplementary Material for \\ EpiMask: Leveraging Epipolar Distance Based Masks in Cross-Attention for Satellite Image Matching}\\[10pt]
    Rahul Deshmukh \quad Aditya Chauhan \quad Avinash Kak\\[1mm]
    {\tt\small deshmuk5@purdue.edu} \quad {\tt\small chauha35@purdue.edu} \quad {\tt\small kak@purdue.edu}\\[1mm]
    Purdue University, West Lafayette
\end{center}

\bigskip
\hrule
\bigskip

\setcounter{section}{0}
\setcounter{subsection}{0}

\renewcommand{\thesection}{\arabic{section}}
\renewcommand{\thesubsection}{\thesection.\arabic{subsection}}

\section{Overview}
In this supplementary material, we provide additional details related to EpiMask. In \cref{sec:architectural_details}, we describe the model architecture and hyperparameters. In \cref{sec:training_strat_ablation_av}, we include an additional ablation study that could not be presented in the main manuscript due to page limitations. In \cref{sec:understanding_model}, we visualize the model's self-attention, cross-attention, and confidence matrices. In \cref{sec:experimental_results}, we present comprehensive quantitative and qualitative results, along with training and validation losses and metrics. Finally, in \cref{sec:ablation_studies}, we provide a fine-grained analysis of all ablation studies.

\section{Architectural Details}\label{sec:architectural_details}

\subsection{Encoder Decoder}
\begin{figure}[H]
    \centering
    \includegraphics[width=\linewidth]{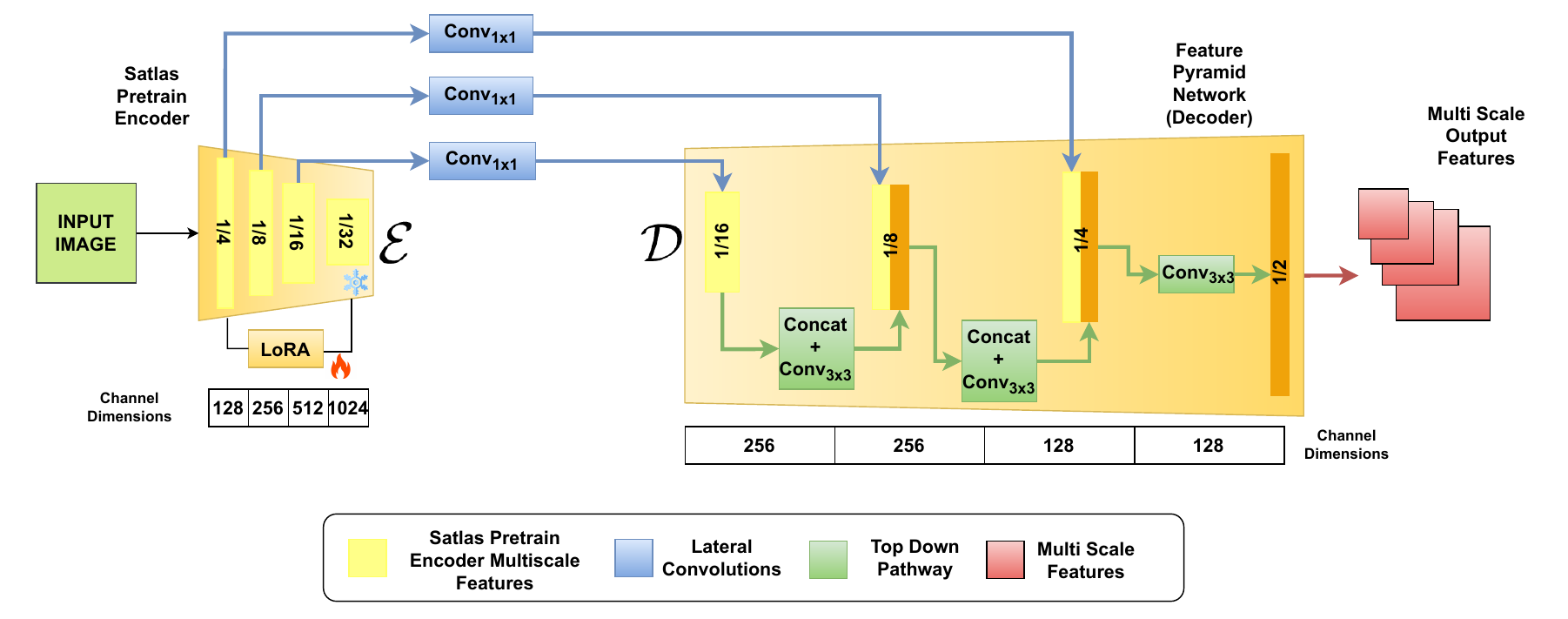}
    \caption{Feature extractor architecture used in EpiMask-HR}
    \label{fig:fpn_decoder}
\end{figure}

Our encoder is based on the Swin Transformer foundation model trained on high-resolution aerial imagery from the Satlas Pretrain dataset \cite{bastani2023satlaspretrain}. The encoder produces multi-scale feature maps at $1/4$, $1/8$, $1/16$, and $1/32$ of the input resolution, with channel dimensions $128$, $256$, $512$, and $1024$, respectively as shown in \cref{fig:fpn_decoder}.

During training, the encoder is kept frozen and adapted using LoRA \cite{hu2022lora} by inserting low-rank adapters to all linear layers of self-attention and reduction modules of the Swin Transformer blocks. We experimented with two configurations, LoRA-16 and LoRA-32. The configuration details are summarized in Table~\ref{tab:lora_configs}.

\begin{table}[h]
    \centering
    \begin{tabular}{ccc}
        \toprule
        \textbf{LoRA Config} & \textbf{Rank (r)} & \textbf{Alpha ($\alpha$)} \\
        \midrule
        LoRA-16 & 16 & 8 \\
        LoRA-32 & 32 & 16 \\
        \bottomrule
    \end{tabular}
    \caption{LoRA configurations used to adapt the Satlas Pretrain encoder.}
    \label{tab:lora_configs}
\end{table}

As shown in \cref{fig:fpn_decoder}, the decoder is a Feature Pyramid Network (FPN) \cite{lin2017feature_FPN}. We take Swin feature maps at $1/32$, $1/16$, $1/8$, and $1/4$ with channel dimensions $\{1024, 512, 256, 128\}$, and apply $1{\times}1$ lateral convolutions to project them to a channel dimensions of the FPN at that level. A top-down pathway upsamples higher-level features and fuses them with lateral features using \textit{concatenation followed by convolution} skip-fusion which carries out concatenation of features along the channel dimension followed by a $3{\times}3$ convolution. This produces FPN outputs at $1/16$, $1/8$, $1/4$, and $1/2$ resolutions with channel dimensions $\{256, 256, 128, 128\}$.

For \textbf{EpiMask-HR} variant, we use the $1/4$ and $1/2$ resolution FPN maps with output channel dimensions $\{d_c{=}128, d_f=128\}$ for $(F_c, F_f)$ respectively. Whereas for the \textbf{EpiMask-LR} variant, we use the $1/8$ and $1/2$ resolution FPN maps with output channel dimension $\{d_c{=}256, d_f{=}128\}$ for $(F_c, F_f)$ respectively.

\subsection{Transformer}
\paragraph{Coarse-Level Masked Attention Transformer:}
The coarse transformer operates on the coarse encoder-decoder output features ($F_c$). For transformer layers in EpiMask-HR, we use a model embedding dimension of $d_c=128$, while for the transformer layers in EpiMask-LR we use $d_c=256$. In both cases we use $N_h^c=8$ attention heads. The architecture consists of $N_c=8$ coarse transformer layers with interleaved linear self-attention and masked cross-attention layers. Each masked cross-attention layer consists of the following sequence of layers: (1) Projection layers for query, key, value; (2) Multi-headed cross-attention with epipolar-distance based masks; (3) Layer-Norm; (4) Feed Forward with two-layer MLP with hidden dimension of $2{*}d_c$ and ReLU activation; (5) Final LayerNorm and residual connections. The input embeddings for the coarse transformer are obtained by flattening the coarse encoder-decoder output feature map ($F_c^L, F_c^R$) into a sequence of shape $[B,\, p^2/r_c^2,\, d_c]$ and adding 2D sinusoidal positional encodings.

For the cross-attention masking, we adopt a warm-up strategy where no masking is applied during the first $N_m = 5$ epochs, after which a linearly decreasing epipolar band width ($b$) is applied across the $N_c$ masked attention layers. The mask band width is linearly decreased from $b=p$ to $b=\gamma p$ across the
masked attention layers. The EpiMask-HR variants are trained with $p=336$ and $\gamma=\{0.4, 0.6\}$, whereas the EpiMask-HR variants are trained with $p=448$ and $\gamma=\{0.4, 0.6\}$.

Finally, the coarse matching confidence matrix is obtained using a dual-softmax operator based matching layer with epipolar-distance based masking. To obtain a set of coarse correspondences, we threshold the confidence matrix with a confidence threshold of $\delta_c=0.3$.

\paragraph{Fine-Level Transformer:}
The fine-level transformer module is the same as LoFTR \cite{Sun_CVPR21_LOFTR} module, which uses a transformer model embedding dimension of $d_f=128$ with $N_h^f=8$ heads and a two-layer transformer self-attention followed by cross-attention. Both self-attention and cross-attention use linear attention \cite{katharopoulos2020transformers_linearattention}. The feed-forward network in transformers has hidden dimension $2 *d_f$ and ReLU activation. Fine-level embeddings are extracted from the higher-resolution FPN feature maps ($F_f$) centered around each coarse match such that for every coarse correspondence, we crop a local $w \times w$ (with $w=5$) window in the fine feature maps. We also concatenate the corresponding coarse features to the fine features within each window before feeding them into the fine-level transformer.

\subsection{Training Hyperparameters}
\paragraph{Image-Patch and Batch Size:}
We follow the SatDepth preprocessing protocol and use $p \times p$ square image patches. For EpiMask-LR we use $p=448$ patches in order to be consistent with SatDepth \cite{deshmukh2025satdepth}. For EpiMask-HR we could not fit a patch size of $p=448$ on our compute and rather use $p=336$ patches. During training, we could fit a batch size of $B=2$ and $B=1$ per gpu for the EpiMask-LR and EpiMask-HR variants respectively.

\paragraph{Optimizer:}
We optimize all trainable parameters using the \texttt{AdamW} optimizer with weight decay $\lambda_w=0.1$ and PyTorch-default betas ($\beta_1{=}0.9, \beta_2{=}0.999$).

\paragraph{Learning-Rate and Warm-Up:}
We use a canonical learning rate of $lr = 8 \times 10^{-3}$ for a reference batch size of $B_{ref}{=}64$, and scale it linearly with the actual batch size. For our configuration this yields $lr_{\text{true}} = 5 \times 10^{-4}$. A linear warm-up schedule is applied for the first $30{,}000$ optimizer steps, during which the learning rate increases from a small fraction ($0.1 * lr_{\text{true}}$) to $lr_{\text{true}}$.

\paragraph{Learning-Rate Scheduler:}
After warm-up we employ a `MultiStep-LR' scheduler at the epoch level. The learning rate is decayed by $\gamma_{lr}=0.5$ at epochs $\{8, 12, 16, 20, 24\}$, and remains constant between milestones. This schedule is used for both EpiMask-HR and EpiMask-LR.

\paragraph{Gradient Accumulation and Clipping:}
We accumulate gradients over $grad\_accum{=}8$ batches before each optimizer step and apply global gradient clipping with a threshold of $\lambda_c=0.5$ to stabilize training, particularly in the early stages when the epipolar-aware transformer layers are still adapting.

\paragraph{Model Size and FLOPs:} We report the total and trainable number of parameters, along with the FLOPs for both the EpiMask-HR and EpiMask-LR variants, in \cref{tab:model_params_and_flops}.

\begin{table}[h]
    \centering
    \begin{tabular}{cccc}
        \toprule
        \textbf{Params / FLOPs} & \textbf{EpiMask-HR} & \textbf{EpiMask-LR} \\
        \midrule
        \textbf{\# Total Params (M)}        & 103   & 107 \\
        \textbf{\# Trainable Params (M)}    & 15.3  & 19.2 \\
        \textbf{FLOPs @ $p{=}336$ (GMACs)}      & 96.35 & 93.21 \\
        \textbf{FLOPs @ $p{=}448$ (GMACs)}      & 174.85 & 165.04 \\
        \bottomrule
    \end{tabular}
    \caption{Model parameter size and FLOPs}
    \label{tab:model_params_and_flops}
\end{table}

\section{Training Strategy Ablation}\label{sec:training_strat_ablation_av}
\begin{figure}[H]
    \centering
    \input{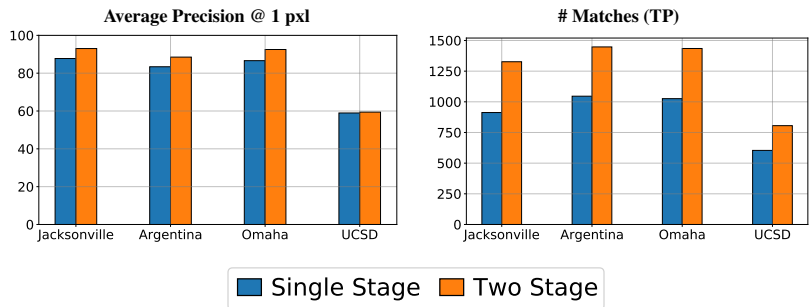}
    \caption{Average precision and number of true-positive matches for all testing AOIs for models trained with single-stage and two-stage training strategies.}
    \label{fig:ablation_performance_training_strategy}
\end{figure}

In the main manuscript, we presented five ablation studies evaluating different components of our model. Here, we include the final ablation study analyzing the impact of our training strategy on overall performance. We compare two approaches: (1) Single-Stage Training and (2) Two-Stage Training. In the single-stage setup, LoRA layers are applied to the pretrained encoder from the beginning, and the entire model (including the decoder and the coarse- and fine-level transformers) is trained jointly from scratch. In contrast, the two-stage setup begins by training the model without LoRA layers to obtain stable weights for the decoder and both transformer modules. In the second stage, we initialize the model with the stage-one weights and introduce learnable LoRA layers into the pretrained encoder for fine-tuning. We present the performance comparison for the two strategies in \cref{fig:ablation_performance_training_strategy}. The two-stage strategy performs better than the single-stage strategy.

\clearpage
\section{Understanding EpiMask}\label{sec:understanding_model}
\begin{figure}[h]
    \centering
    \input{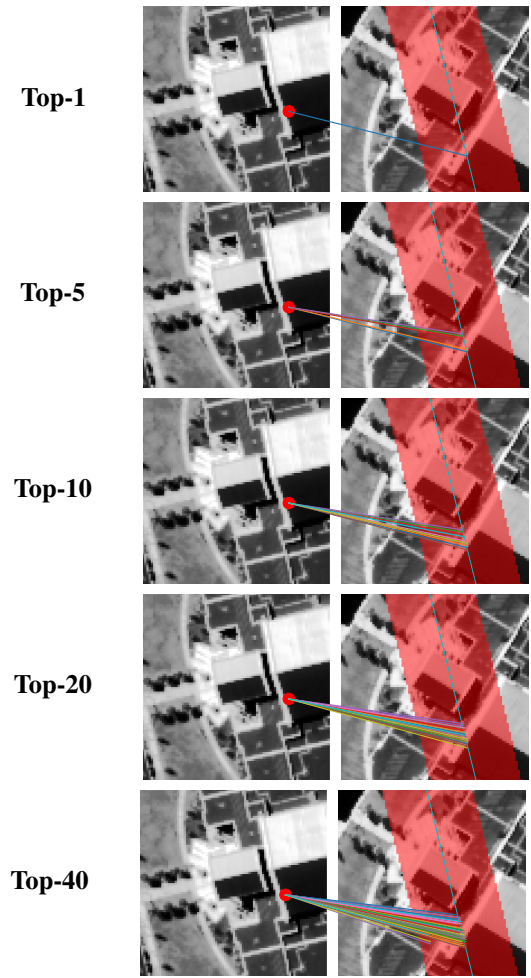}
    \caption{Visualization of Top-K confidence scores in the masked coarse matching module. Shown are Top-K matching points (selected by sorting the confidence scores and taking the K highest) in the right image patch corresponding to the query pixel in the left image patch. In each figure, the query pixel is shown as red dot in the left image patch with corresponding epipolar line (cyan line) and mask (red band) in the right image patch. Top-K matching locations to the query pixel are displayed using lines originating from the query pixel. We observe that the model assigns the highest matching score (Top-1) to the true correspondence, while the remaining high-confidence candidates (up to Top-20) lie along the epipolar line as reasonable alternatives.}
    \label{fig:conf_mat_plots}
\end{figure}

\begin{figure}[h]
    \centering
    \input{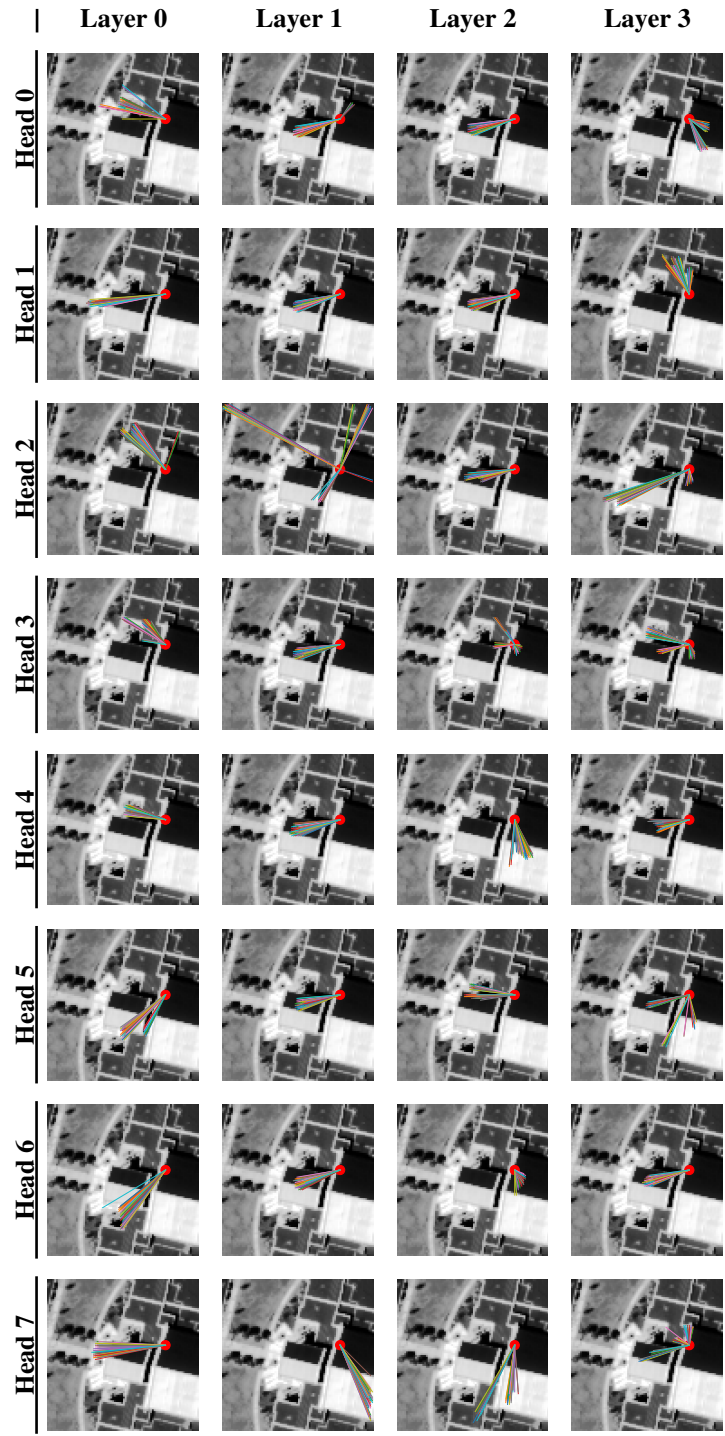}
    \caption{Visualization of self-attention maps in the coarse transformer. Shown are attention maps for each self-attention layer and individual heads within the multi-headed attention. In each figure, the query pixel is shown as red dot and top 40 attention locations are displayed using lines originating from the query pixel. We observe that different attention heads in the self-attention layers exhibit different attentional behaviors: While some attend locally (\eg Layer-0 and Head-0) others capture long-range dependencies (\eg Layer-1 and Head-2).}
    \label{fig:self_attn_masks_plots}
\end{figure}

\begin{figure}[h]
    \centering
    \input{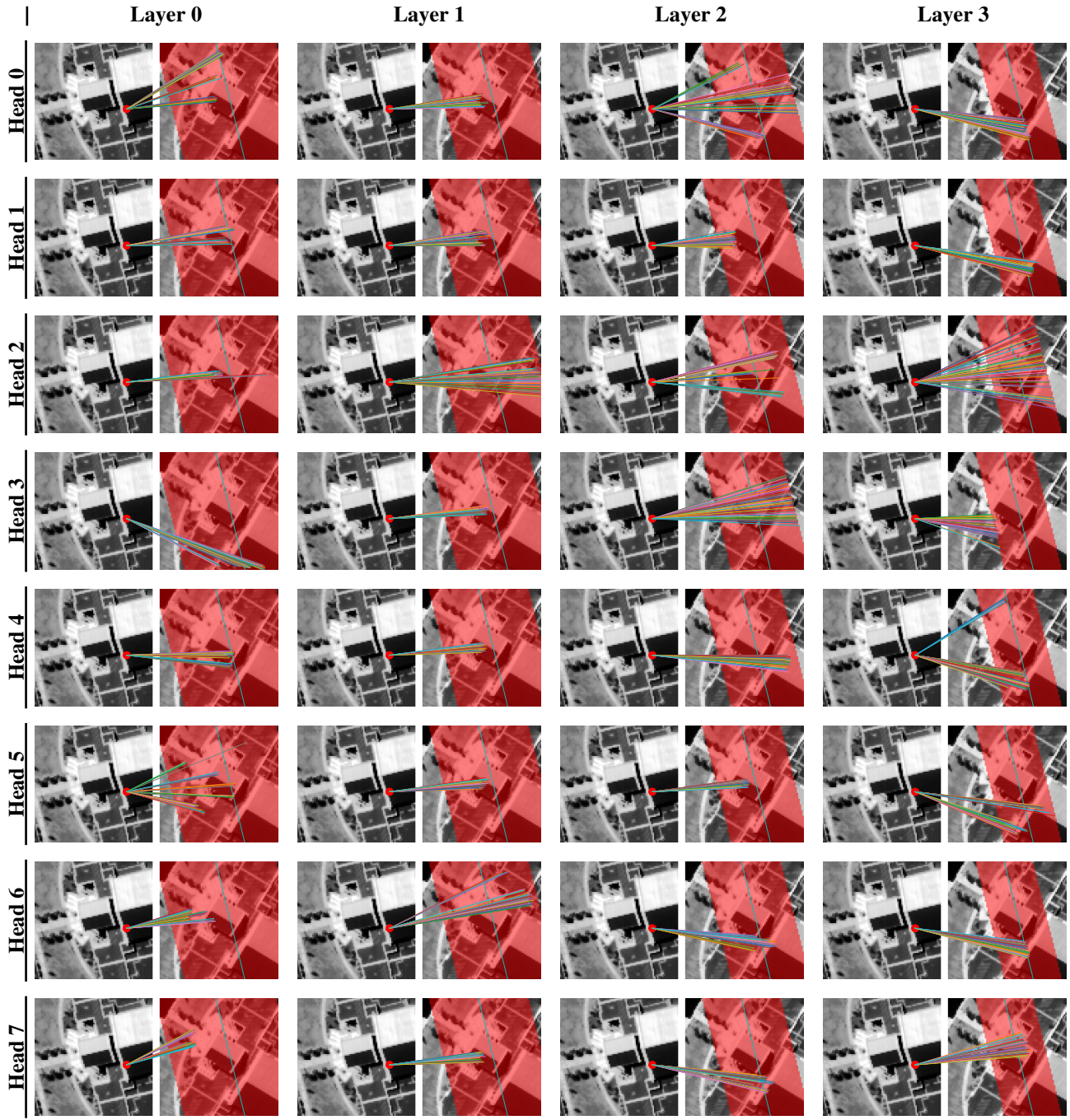}
    \caption{Visualization of masked-cross-attention maps in the coarse transformer. Shown are attention maps for each masked cross-attention layer and individual heads within the multi-headed attention. In each figure, the query pixel is shown as red dot in the left image patch with corresponding epipolar line (cyan line) and mask (red band) in the right image patch. Top 40 attention locations corresponding to the query pixel are displayed using lines originating from the query pixel. We observe that the earlier layers often do not focus on the true corresponding region. However, by the final layer (Layer-3), five out of eight heads (Heads-0,1,4,5,6) attend to the correct region.}
    \label{fig:cross_attn_masks_plots}
\end{figure}

To better understand the internal workings of EpiMask, we visualize the self-attention maps (\cref{fig:self_attn_masks_plots}) and masked cross-attention maps (\cref{fig:cross_attn_masks_plots}) across all layers and heads of our coarse-level masked attention transformer. We also show the coarse-level matching confidence matrix in \cref{fig:conf_mat_plots}.

From \cref{fig:self_attn_masks_plots}, we observe that different attention heads across the self-attention layers specialize in diverse pattern, some attend locally, while others capture long-range dependencies. In the case of masked cross-attention (\cref{fig:cross_attn_masks_plots}), earlier layers often do not focus on the true corresponding region. However, by the final layer, five out of eight heads converge to the correct region, indicating progressive refinement. Finally, the matching confidence visualization (\cref{fig:conf_mat_plots}) shows that the model assigns the highest matching score (Top-1) to the true correspondence, while the remaining high-confidence candidates (up to Top-20) lie along the epipolar line as reasonable alternatives.

\clearpage
\section{Experimental Results}\label{sec:experimental_results}
\begin{figure}[h]
    \centering
    \input{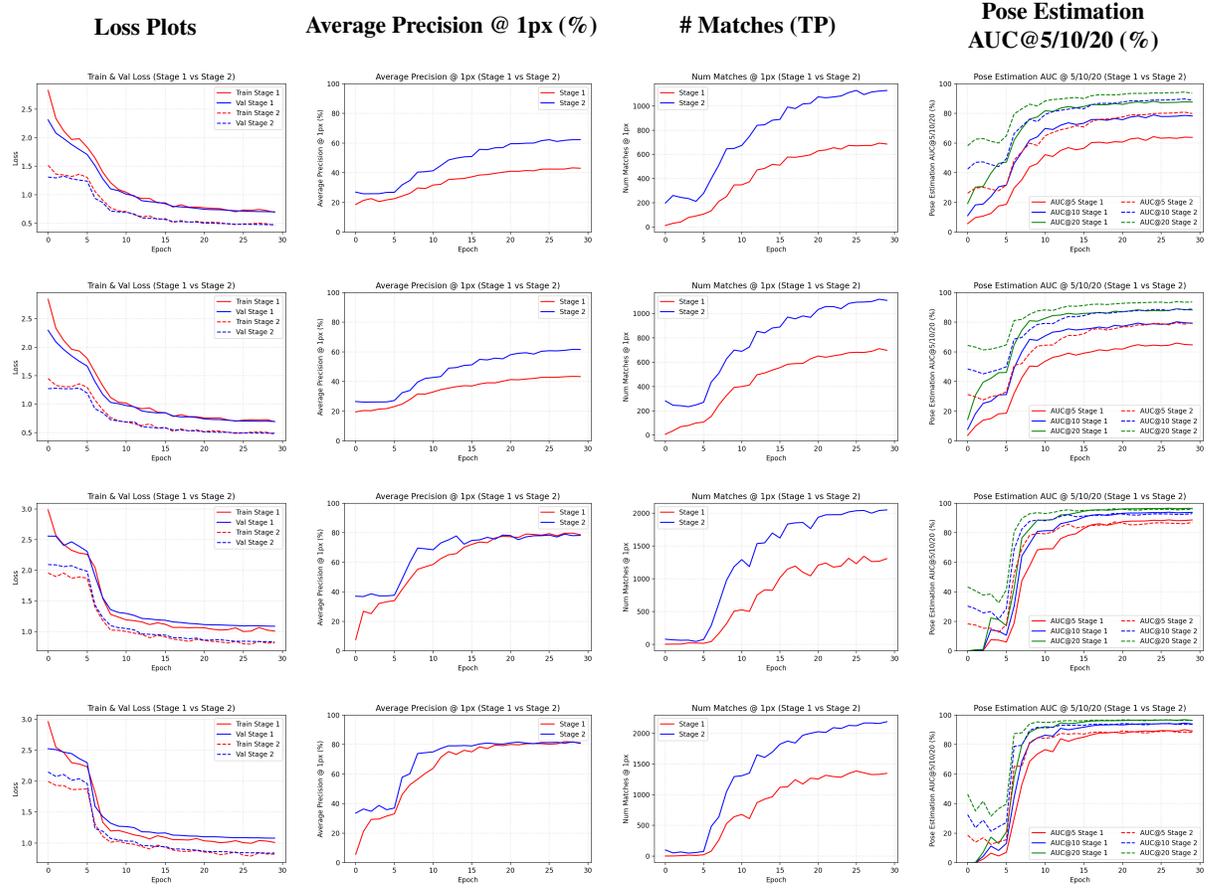}
    \caption{Plots for loss and metrics for our four model configurations. The first column shows the training and validation loss per epoch for both stages of training. The other columns show metrics (Precision, Number of true-positive matches, and pose estimation AUC) per epoch on the validation set for both stages of training.}
    \label{fig:tf_plots}
\end{figure}

In the main manuscript, due to page limitations we presented quantitative results only for Jacksonville and San Fernando testing AOIs. Here we present comprehensive quantitative aggregated results for each testing AOI in \cref{fig:matcher_performance_all}. We also present the quantitative results for `Simulated-Rotation' experiment for all testing AOIs in \cref{fig:all_model_comparison_plots_for4Aois}. Following SatDepth ~\cite{deshmukh2025satdepth} we present qualitative results for large view-angle and time differences in \cref{fig:qualitative_matching_plots_view_angle,fig:qualitative_matching_plots_time_diff} respectively. Finally, we present the plots for training and validation losses and metrics for our four model configurations in \cref{fig:tf_plots}.

\begin{figure}[h]
    \centering
    \input{tables/all_aois_matcher_performance_table.tex}
    \caption{Weighted average of Precision, Pose error, and number of True Positive (TP) matches over all testing image patches for all testing AOIs of SatDepth.}
    \label{fig:matcher_performance_all}
\end{figure}

\begin{figure}[h]
    \centering
    \input{./figure_tables/All4models_performance_comparison.tex}
    \caption{Comparison of all model average precision and number of true-positive matches for all testing AOIs of SatDepth in the simulated rotation experiment. EpiMask consistently achieves the highest precision and detects the largest number of matches w.r.t varying view-angle differences $\alpha^{v}$ and track-angle differences $\alpha^{t}$.}
    \label{fig:all_model_comparison_plots_for4Aois}
\end{figure}

\clearpage
\begin{figure}[H]
    \centering
    \input{./figure_tables/hard_pairs_intersection_matching_plots_table.tex}
    \caption{  
    Qualitative comparison of our model results against other models for large view-angle difference ($\alpha^v$) -- our model has the highest precision score. Precision (P) and number of matches (N) are displayed at the top of each plot. Image pair names, time difference ($\Delta t$), view-angle difference ($\alpha^v$), and track-angle difference ($\alpha^t$) are displayed at the bottom. The green lines depict 40 randomly chosen true matches.}
    \label{fig:qualitative_matching_plots_view_angle}
\end{figure}

\begin{figure}[H]
    \centering
    \input{./figure_tables/hard_pairs_time_diff_matching_plots_table.tex}
    \caption{  
    Qualitative comparison of our model results against other models for large time difference ($\Delta t$) -- our model has the highest precision score. Precision (P) and number of matches (N) are displayed at the top of each plot. Image pair names, time difference ($\Delta t$), view-angle difference ($\alpha^v$), and track-angle difference ($\alpha^t$) are displayed at the bottom. The green lines depict 40 randomly chosen true matches.}
    \label{fig:qualitative_matching_plots_time_diff}
\end{figure}

\clearpage
\section{Ablation Studies}\label{sec:ablation_studies}
In the main manuscript we presented averaged results for each testing AOIs, where the average was over all view and track angles. In the following sub-sections we present a fine-grained analysis for each ablation study w.r.t. different ranges for view angle and track angle differences.

\subsection{Resolution Ablation}
\begin{figure}[h]
    \centering
    \input{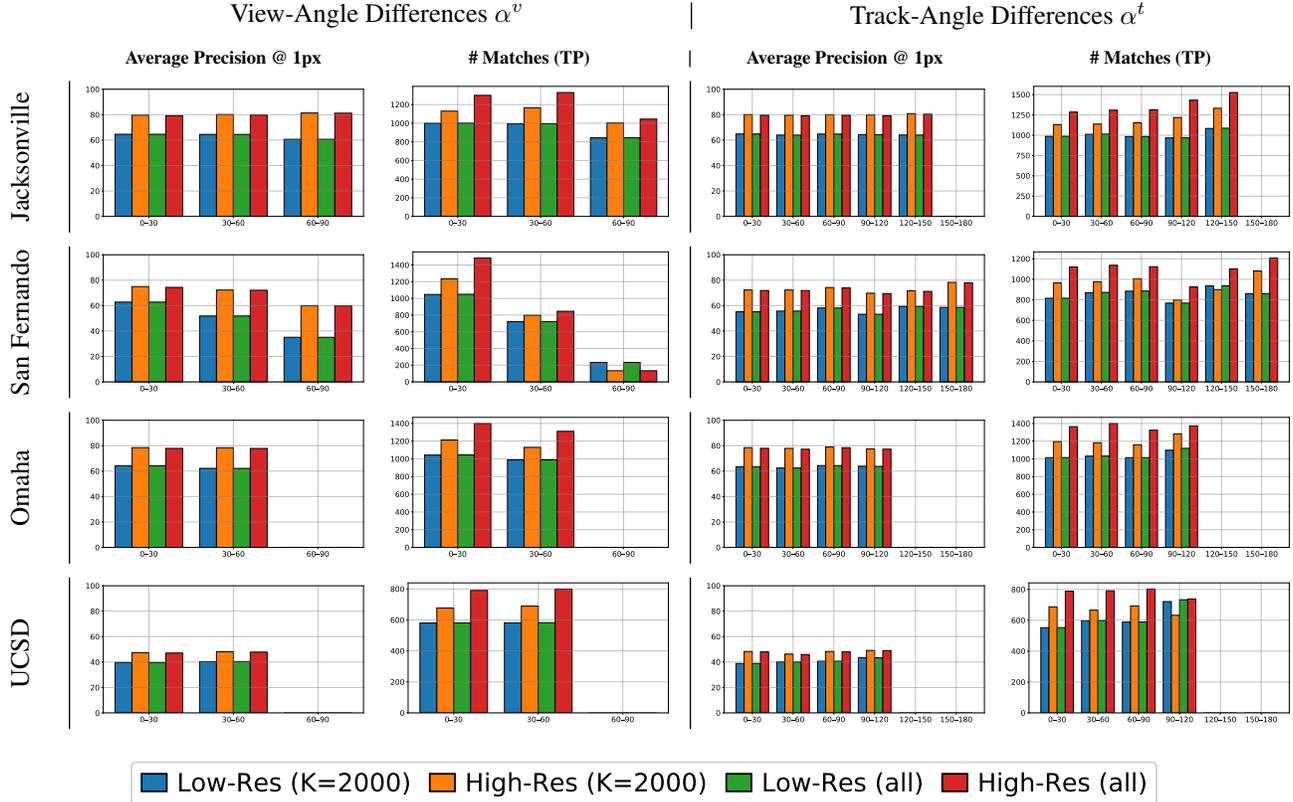}
    \caption{Average precision and number of true positive matches for resolution ablation.}
    \label{fig:resolution_ablation_plots}
\end{figure}

In the main manuscript we presented results for the resolution ablation for all testing AOIs averaged over all angles. In this section we present a fine-grained analysis over different ranges of view angle ($\alpha^v$) and track angle differences ($\alpha^t$) (see \cref{fig:resolution_ablation_plots}). The High-Res configuration performs the best. 

\clearpage
\subsection{Attention Mask Width Ablation}
\begin{figure}[h]
    \centering
    \input{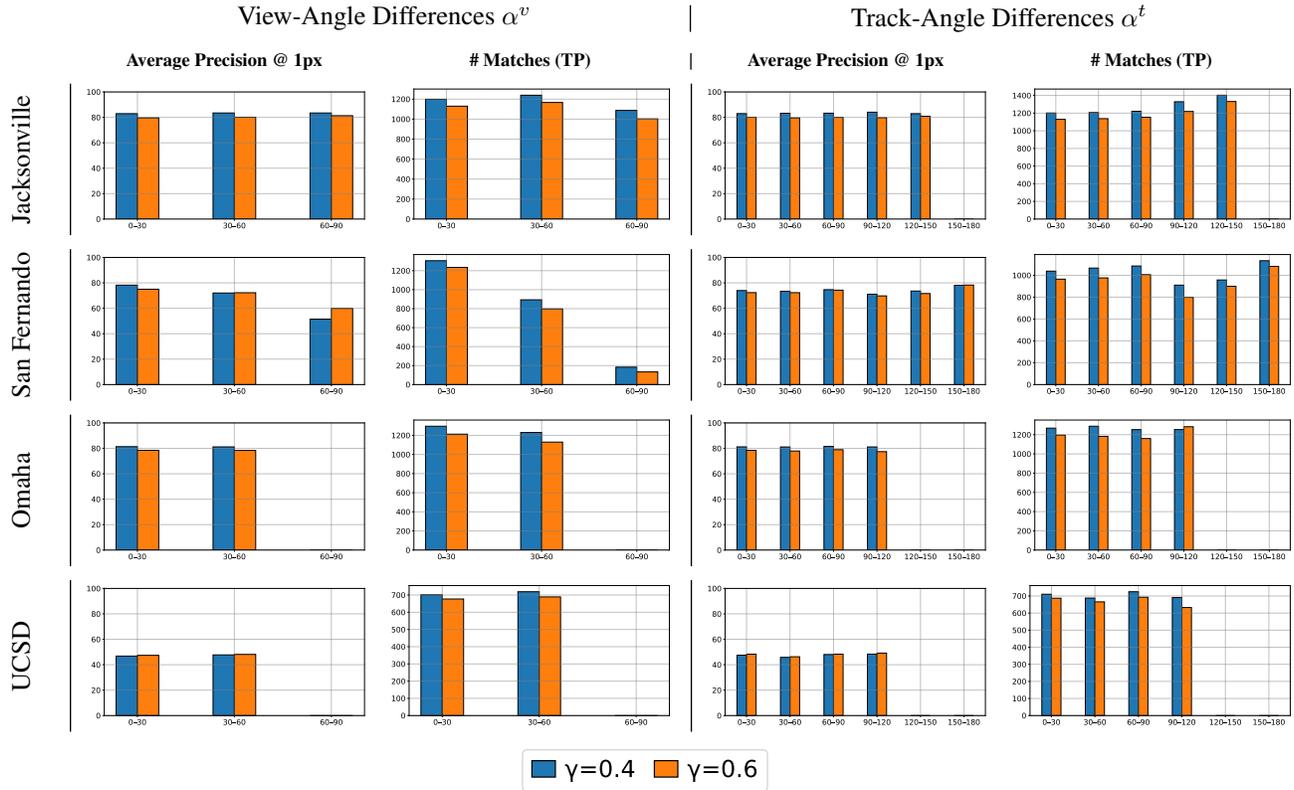}
    \caption{Average precision and number of true positive matches for attention mask width ablation.}
    \label{fig:attn_mask_width_ablation_plots}
\end{figure}

In the main manuscript we presented results for the attention mask width ablation for all testing AOIs averaged over all angles. In this section we present a fine-grained analysis over different ranges of view angle ($\alpha^v$) and track angle differences ($\alpha^t$) (see \cref{fig:attn_mask_width_ablation_plots}). As shown in \cref{fig:attn_mask_width_ablation_plots}, performance remains largely unchanged, indicating that the model inherently focuses on geometrically consistent regions within the epipolar band. This robustness suggests that fine-grained tuning of mask width is unnecessary. In practice, narrower masks may be preferred for newer satellites with accurate pose metadata, while wider masks can better handle older sensors with noisier estimates of camera pose.

\clearpage
\subsection{Positional Encoding Ablation}
\begin{figure}[h]
    \centering
    \input{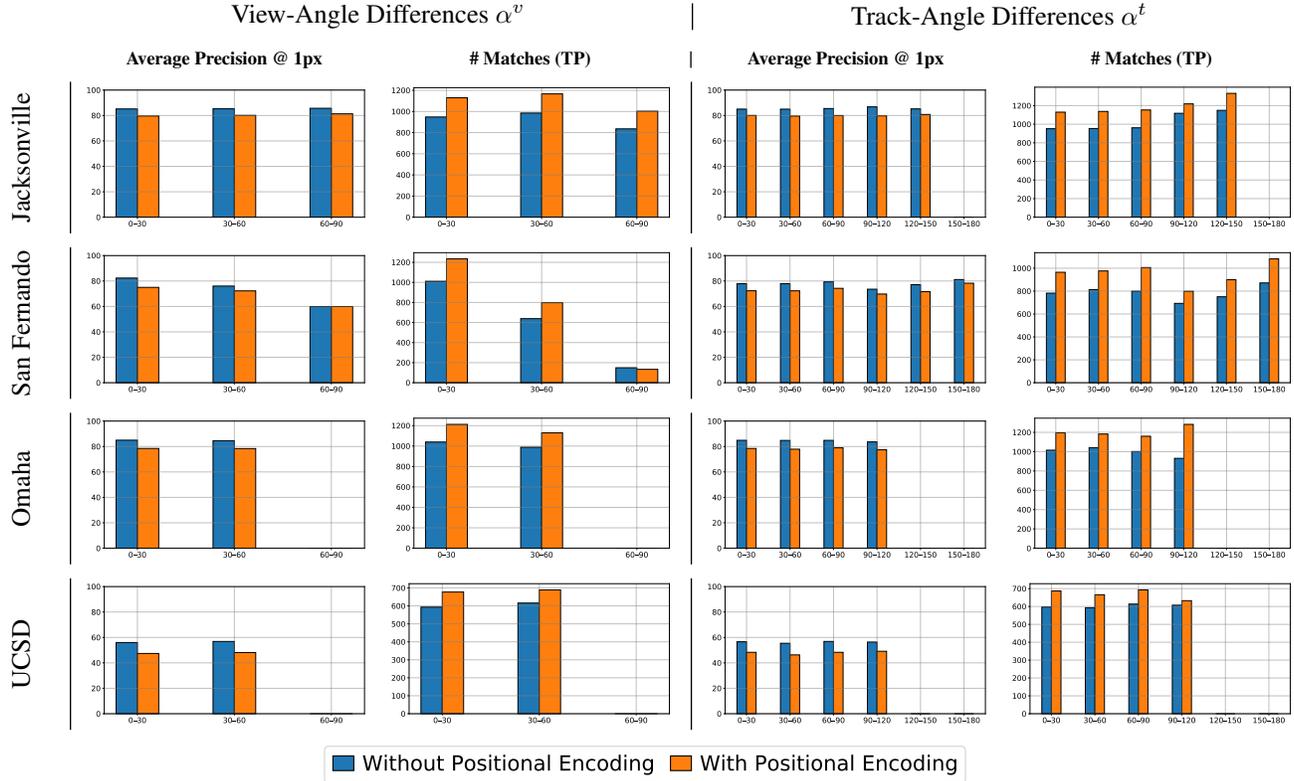}
    \caption{Average precision and number of true positive matches for positional encoding ablation.}
    \label{fig:positional_encoding_ablation_plots}
\end{figure}

In the main manuscript we presented results for the positional encoding ablation for all testing AOIs averaged over all angles. In this section we present a fine-grained analysis over different ranges of view angle ($\alpha^v$) and track angle differences ($\alpha^t$) (see \cref{fig:positional_encoding_ablation_plots}). As shown in \cref{fig:positional_encoding_ablation_plots}, average precision remains similar, but positional encodings consistently increase true-positive matches. 

\clearpage
\subsection{Fine-Tuning Ablation}
\begin{figure}[h]
    \centering
    \input{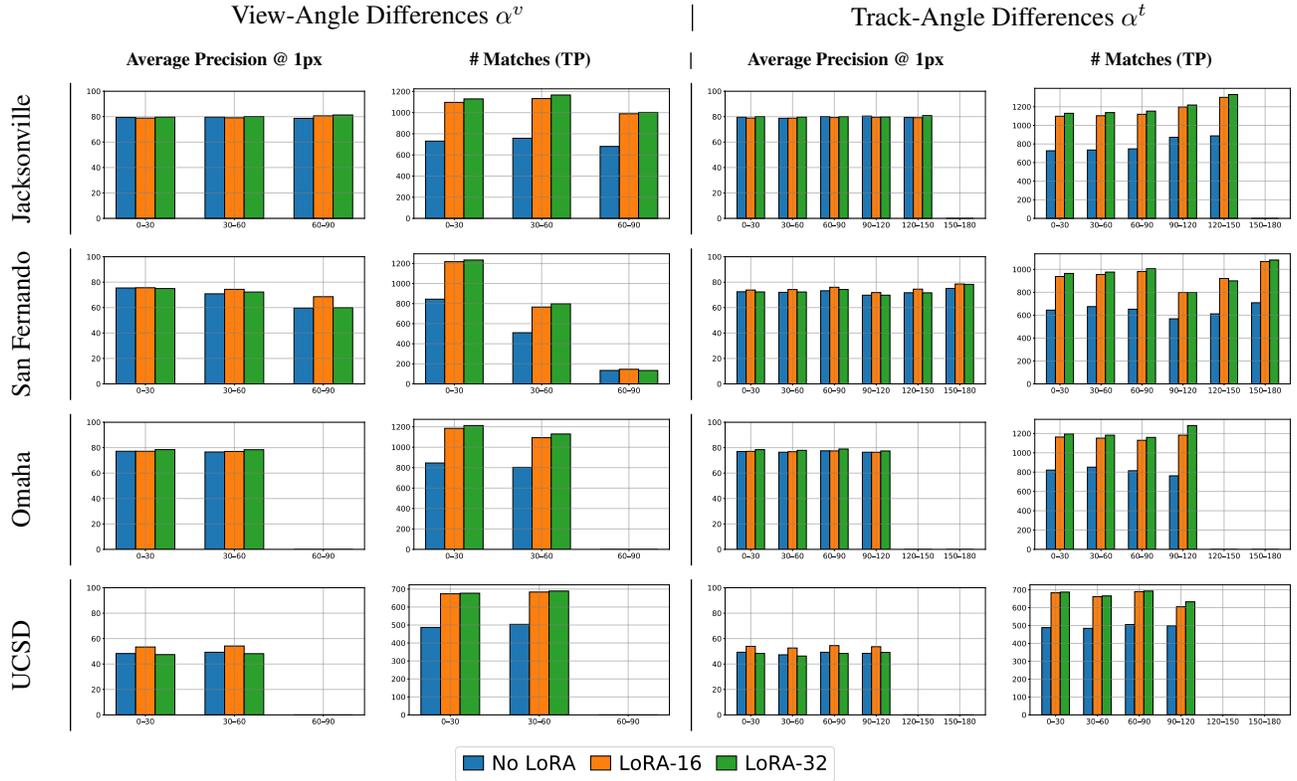}
    \caption{Average precision and number of true positive matches for fine-tuning ablation.}
    \label{fig:fine_tuning_ablation_plots}
\end{figure}

In the main manuscript we presented results for the fine-tuning ablation for all testing AOIs averaged over all angles. In this section we present a fine-grained analysis over different ranges of view angle ($\alpha^v$) and track angle differences ($\alpha^t$) (see \cref{fig:fine_tuning_ablation_plots}). As shown in \cref{fig:fine_tuning_ablation_plots}, LoRA significantly increases the number of true positives, indicating that lightweight fine-tuning effectively adapts the pretrained backbone to our matching task. Increasing the rank from 16 to 32 provides negligible gains, suggesting that a moderate rank of 16 strikes a good balance between parameter efficiency and adaptation quality.

\clearpage
\subsection{Feature Extractor Ablation}
\begin{figure}[h]
    \centering
    \input{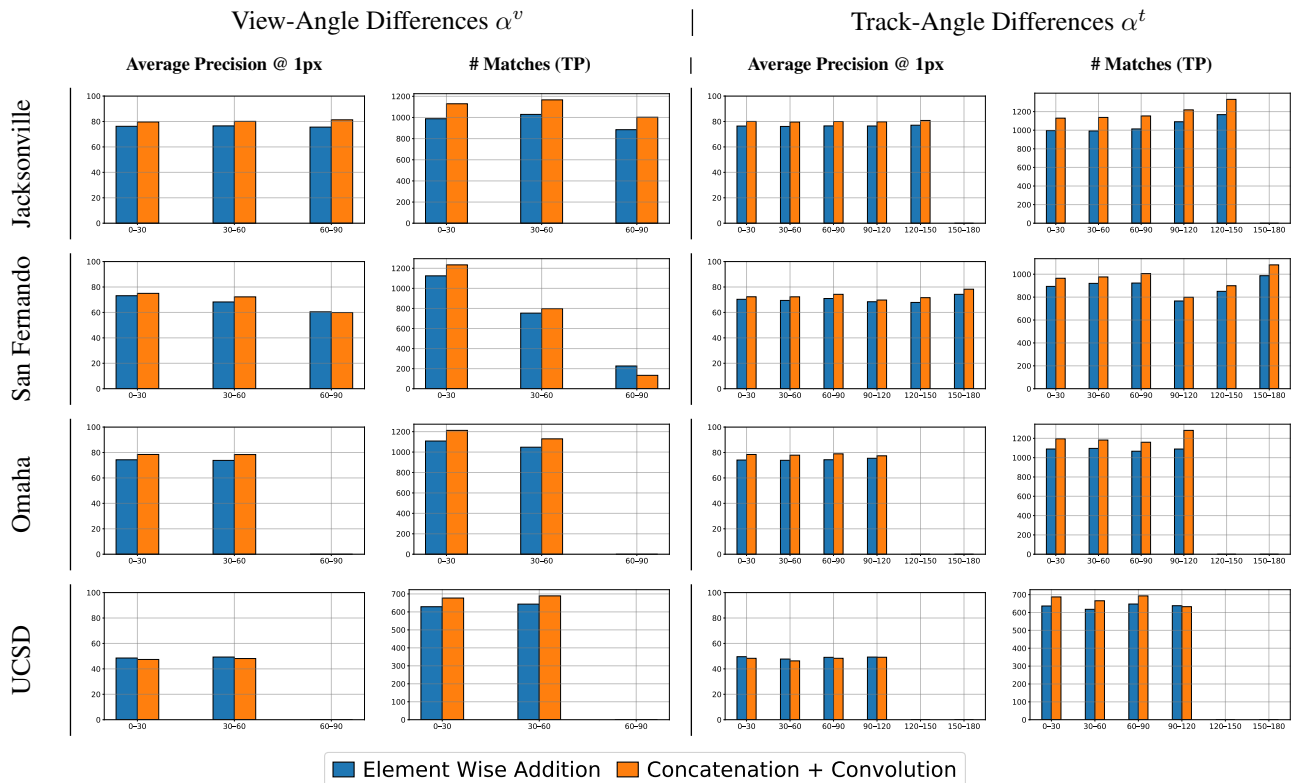}
    \caption{Average precision and number of true positive matches for feature extractor ablation.}
    \label{fig:feature_extraction_ablation_plots}
\end{figure}

In the main manuscript we presented results for the feature extractor ablation for all testing AOIs averaged over all angles. In this section we present a fine-grained analysis over different ranges of view angle ($\alpha^v$) and track angle differences ($\alpha^t$) (see \cref{fig:feature_extraction_ablation_plots}). As shown in \cref{fig:feature_extraction_ablation_plots}, the concatenation followed by convolution strategy performs better compared to na\"{\i}ve element-wise addition.

\clearpage
\subsection{Training Strategy Ablation}
\begin{figure}[h]
    \centering
    \input{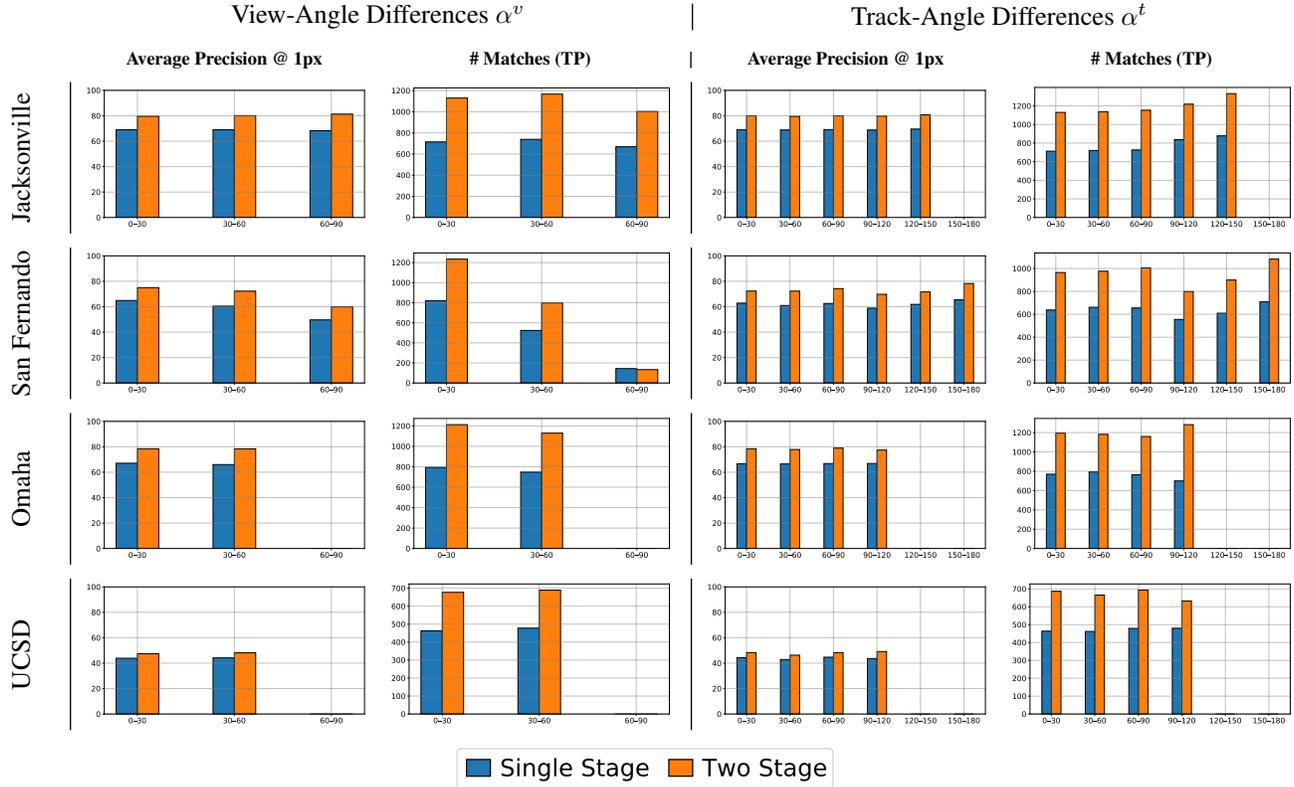}
    \caption{Average precision and number of true positive matches for training strategy ablation.}
    \label{fig:training_strategy_ablation_plots}
\end{figure}

Earlier in \cref{sec:training_strat_ablation_av} we presented results for the training strategy ablation for all testing AOIs averaged over all angles. In this section we present a fine-grained analysis over different ranges of view angle ($\alpha^v$) and track angle differences ($\alpha^t$) (see \cref{fig:training_strategy_ablation_plots}). As shown in \cref{fig:training_strategy_ablation_plots}, the two-stage training strategy performs better than the single stage.

{
    \small
    \bibliographystyle{ieeenat_fullname}
    \bibliography{main}
}


\end{document}